\pdfoutput=1

\documentclass[11pt]{article}

\usepackage[final]{acl}
\usepackage{graphicx}
\usepackage{tabularx} 

\usepackage{multirow}   
\usepackage{xcolor}     
\usepackage{colortbl}
\usepackage{amsmath,amsfonts}
\usepackage{algorithmic}
\usepackage{adjustbox}
\usepackage{algorithm}
\usepackage{array}
\usepackage{subfigure}
\usepackage{subcaption}
\usepackage{textcomp}
\usepackage{stfloats}
\usepackage{url}
\usepackage{verbatim}
\usepackage{graphicx}
\usepackage{booktabs}
\usepackage{graphicx}%
\usepackage{float}
\usepackage{caption}
\usepackage{times}
\usepackage{latexsym}
\usepackage{natbib}
\usepackage{balance}
\usepackage[normalem]{ulem}

\definecolor{lightgreen}{RGB}{173, 216, 230}
\definecolor{lightred}{RGB}{255, 218, 185}
\definecolor{lightpurple}{RGB}{216, 191, 216} 
\definecolor{lightpinkred}{RGB}{255, 182, 193}

\usepackage{makecell}
\usepackage{caption}
\usepackage{xurl}
\usepackage{amsmath}
\usepackage[table]{xcolor}  
\usepackage{soul}

\usepackage{array}
\usepackage{geometry}

\usepackage{tikz}
\usepackage{enumitem}
\usepackage{arydshln}

\usepackage[T1]{fontenc}

\usepackage[utf8]{inputenc}

\usepackage{microtype}

\usepackage{inconsolata}

\usepackage{graphicx}

\def\method{KAIR}

\title{Finding What Matters: Anchoring Context Knowledge with Evolving Indices for Iterative Retrieval}

\author{Mingyan Wu$^{1}$\footnotemark[1], Zhenghao Liu$^{1}$\thanks{\ \ indicates equal contribution.}\thanks{ \ \ indicates corresponding author.}, Xinze Li$^{1}$, Yuqing Lan$^{1}$, Yukun Yan$^{2}$\footnotemark[2],\\ \textbf{Shuo Wang$^{2}$, Cheng Yang$^{3}$, Minghe Yu$^{1}$, Zheni Zeng$^{2}$, Maosong Sun$^{2}$} \\ 
$^1$School of Computer Science and Engineering, Northeastern University, China \\
$^2$Department of Computer Science and Technology, Institute for AI, Tsinghua University, China \\
Beijing National Research Center for Information Science and Technology, China \\
$^3$School of Computer Science, Beijing University of Posts and Telecommunications, China
}

\begin{document}
\maketitle

\begin{abstract}
Retrieval-Augmented Generation (RAG) has become a dominant paradigm for mitigating hallucinations in Large Language Models (LLMs) by incorporating external knowledge. However, existing RAG systems often struggle to effectively integrate and reason over key evidence scattered across noisy retrieved documents, particularly in multi-hop scenarios.
In this paper, we propose \textbf{\method{}}, a \textbf{K}nowledge \textbf{A}nchoring framework for \textbf{I}terative \textbf{R}etrieval that anchors knowledge within retrieved knowledge to guide LLMs to locate the key information. During iterative retrieval, \method{} progressively updates the knowledge index to anchor salient evidence from retrieved documents. The evolving index serves as a navigational anchoring index that enables the LLM to assess knowledge sufficiency and formulate subsequent retrieval queries. Finally, \method{} generates answers by jointly leveraging the retrieved documents and the finalized anchoring index.
Experiments on four multi-hop question answering benchmarks demonstrate that \method{} consistently outperforms strong RAG baselines. Further analysis shows that \method{} effectively anchors key knowledge and alleviates the context noise during iterative retrieval, improving the LLM's ability to associate and reason over dispersed evidence across retrieved documents. 
All code and data are available at \url{https://github.com/NEUIR/KAIR}.
\end{abstract}

\section{Introduction}
Retrieval-Augmented Generation (RAG) models typically retrieve relevant passages and incorporate them as contextual inputs, thereby helping Large Language Models (LLMs) to generate accurate and grounded responses~\cite{ram2023context, lewis2020retrieval, shi2023replug, shuster2021retrieval}. To handle multi-hop question answering scenarios where evidence is frequently distributed across multiple documents~\cite{li2024structrag}, several studies have proposed iterative retrieval strategies that progressively seek and accumulate relevant knowledge to support answer generation~\citep{jin2025search, li2025search, kuratov2024babilong}. Nevertheless, as the number of retrieval steps increases, these RAG systems inevitably introduce more noisy or irrelevant information. The model frequently fails to identify and connect the fragmented evidence, which can substantially degrade the quality of the generated outputs~\cite{cuconasu2024power}. 

\begin{figure}[t] 
\centering
    \includegraphics[width=0.48\textwidth]{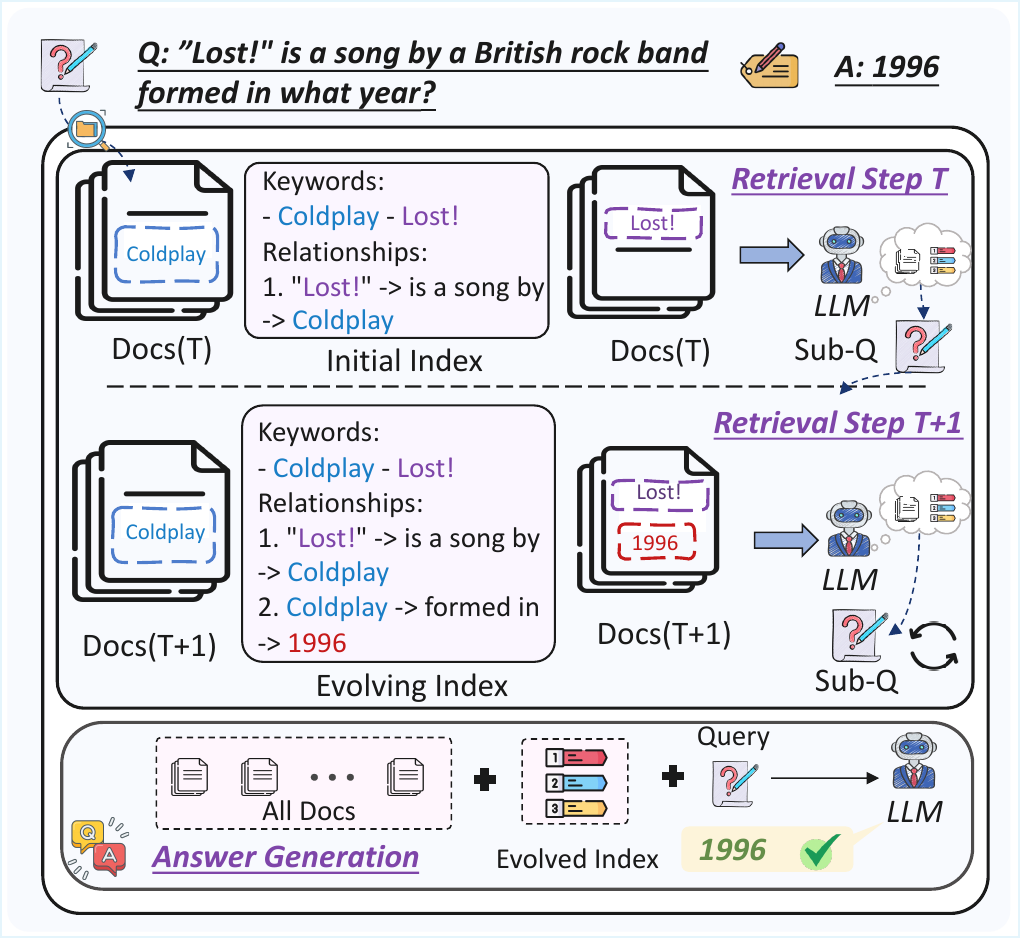}
    \caption{Illustration of \method{}. The index is progressively updated for anchoring context knowledge.} \label{fig:intro_fig}
\end{figure}

RAG performance varies dramatically depending on the position of the golden evidence within the retrieved evidence list, due to the ``lost in the middle'' phenomenon~\cite{liu2023lost, DBLP:conf/iclr/JinY0A25}. This performance variance is further amplified under deep retrieval, which has motivated several studies to focus on refining retrieved evidence by filtering out noise through reranking or summarization at each retrieval step~\cite{wang2025deepnotenotecentricdeepretrievalaugmented, trivedi2023interleaving}. However, these methods often fail to associate knowledge across different retrieval steps and frequently lose essential information needed for LLMs to answer the given questions. Recently, several studies \cite{edge2024local, guo2024lightrag, DBLP:conf/icml/GutierrezSQZ025} have organized extracted entities and facts into graph structures to improve knowledge representation or chunk localization~\cite{DBLP:conf/acl/FangMM25}. However, whether such structures can further serve as dynamic navigational guides to help LLMs progressively consolidate knowledge across multiple retrieval iterations remains largely underexplored. To address this limitation, we shift our focus from knowledge refinement to knowledge anchoring, which preserves all retrieved evidence in the input context and guides LLMs to identify and leverage key information from the retrieved evidence. As shown in Figure~\ref{fig:intro_fig}, an index provides a well-defined navigational guidance for explicitly modeling keywords and their relations. Such indices can be incrementally updated by introducing newly identified keywords, for example, adding ``1996'' and linking it with ``Coldplay'' through the ``formed in'' relation. This index updating process progressively anchors the retrieved documents, enabling more effective knowledge utilization.

In this paper, we propose \method{}, a novel Knowledge Anchoring framework that guides LLMs to capture and consolidate knowledge from deep retrieval via evolving anchoring indices. During the iterative retrieval process, \method{} progressively updates an in-context anchoring index to explicitly anchor salient evidence extracted from newly retrieved documents. Rather than simply accumulating text, this evolving index acts as a navigational guide, teaching LLMs to associate and integrate relevant knowledge from the input context even as more noise is introduced with subsequent retrieval steps. At each retrieval iteration, \method{} leverages the index to assess whether the current knowledge is sufficient to answer the original query. If the knowledge is insufficient, the model uses the index to formulate targeted queries for the next retrieval step. Once the knowledge is deemed adequate, the final response is generated by jointly leveraging the raw aggregated documents and the finalized knowledge index.

Our experiments on four multi-hop QA datasets demonstrate the effectiveness of \method{}. Rather than refining knowledge for question answering, \method{} focuses on knowledge anchoring and improves both the retrieval and QA modules within an iterative RAG framework. Further analysis shows that, as the retrieval process progresses, the index anchors an increasing number of related key pieces of information into the index, so as to alleviate the context noise, and subsequently modulates the attention mechanism of LLMs to aggregate distributed evidence from the retrieved documents, enabling more robust and accurate performance.

\section{Related Work}
Retrieval-Augmented Generation (RAG) enhances Large Language Models (LLMs) by incorporating retrieved passages into the context window, facilitating the generation of more accurate outputs~\cite{ram2023context, lewis2020retrieval, wu2025rankcot, shuster2021retrieval}. However, conventional RAG approaches often suffer from insufficient knowledge acquisition, particularly in multi-hop question answering tasks that require integrating evidence scattered across multiple passages.

To address this limitation, several studies enhance retrieval interactions to acquire sufficient knowledge for answering multi-hop questions~\cite{DBLP:conf/acl/SuTA0024, DBLP:conf/emnlp/JiangXGSLDYCN23}. One line of work employs iterative retrieval strategies in RAG systems, progressively aggregating the information required for reasoning~\cite{shao2023enhancing, trivedi2023interleaving}. Another line of work leverages the reasoning capabilities of LLMs to invoke retrieval during the reasoning process, enabling models to adaptively use retrieval tools whenever additional knowledge is needed to answer a given question~\cite{li2025search, jin2025search}. However, as retrieval depth increases, the accumulated context inevitably contains more redundant and irrelevant passages~\cite{DBLP:conf/iclr/JinY0A25, liu2023lost}. As a result, LLMs often struggle to effectively identify query-relevant evidence from noisy retrieval results, substantially limiting knowledge utilization in multi-hop question answering scenarios.

To mitigate this issue, prior studies have primarily focused on reducing retrieval noise through passage reranking~\cite{asai2024selfrag, wu2025rankcot} or salient information summarization~\cite{vig2021exploring, wang2025deepnotenotecentricdeepretrievalaugmented}. However, these approaches often inevitably introduce information loss during knowledge refinement and fail to effectively capture the relationships among retrieved evidence.
Rather than refining retrieved passages directly, several recent works~\cite{edge2024local, guo2024lightrag, DBLP:conf/icml/GutierrezSQZ025} construct graph structures over passages to support LLM-based question answering. Specifically, these methods extract key entities and facts from different chunks or passages and organize them into a graph structure~\cite{DBLP:conf/acl/FangMM25}. These methods are typically designed either to represent knowledge in a more structured manner or to facilitate the localization of query-relevant chunks. Nevertheless, whether such structures can further serve as navigational structures that help LLMs progressively consolidate knowledge across multiple retrieval iterations remains largely underexplored.

\section{Methodology}
\begin{figure*}[t] 
\centering
    \includegraphics[width=1.0\textwidth]{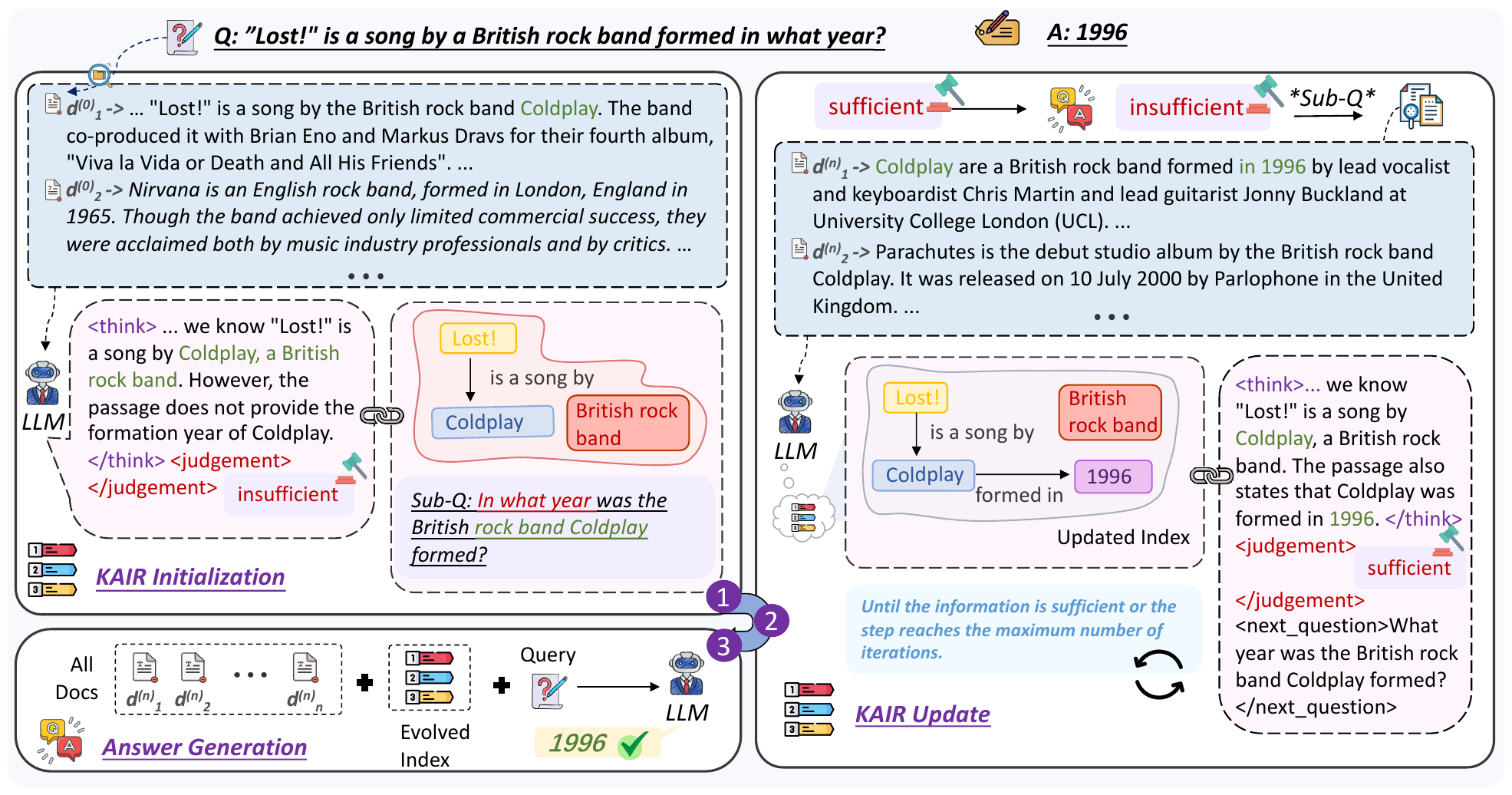}
    \caption{Illustration of \method{}.} \label{fig:main_fig}
\end{figure*}

As illustrated in Figure~\ref{fig:main_fig}, this section presents the proposed \method{}, which introduces an evolving in-context index to anchor retrieved knowledge within an iterative RAG framework. We first describe the overall architecture of \method{}, highlighting how the evolving index is integrated to support reasoning, query generation, and question answering (Sec.~\ref{section:preliminary}). We then elaborate on the construction and dynamic updating of the in-context index, which functions as a context-aware knowledge index to enhance the knowledge association and comprehension capabilities of LLMs (Sec.~\ref{section:index}).

\subsection{Knowledge Anchoring for Iterative RAG}
\label{section:preliminary}
Given an initial query $q_0$ that requires multi-hop retrieval, existing approaches~\cite{shao2023enhancing,trivedi2023interleaving} typically perform iterative retrieval to progressively accumulate evidence for answering the query. To better anchor these retrieved pieces of evidence in context, \method{} constructs an evolving index to guide LLMs in effectively utilizing the knowledge contained in the retrieved documents.

\textbf{Iterative Retrieval.} At the $t$-th retrieval step, the framework retrieves a set of relevant documents $D_t$ conditioned on the previously refined query $q_{t-1}$:
\begin{equation}
D_t = \text{Retriever}(q_{t-1}),
\end{equation}
where $1 \leq t \leq T$ and $T$ denotes the maximum number of retrieval iterations. 
After completing the iterative retrieval process, all retrieved documents are aggregated to form the complete document set:
\begin{equation}\label{eq:all_document}
    \mathcal{D} = D_1 \cup D_2 \cup \dots \cup D_k,
\end{equation}
where $k$ denotes the actual number of retrieval steps performed. The LLM then generates the answer $y$ to the original query $q_0$ based on the aggregated document set $\mathcal{D}$:
\begin{equation}
    y = \mathcal{LLM}(q_0, \mathcal{D}).
\end{equation}
As the number of retrieval iterations increases, $\mathcal{D}$ may accumulate noisy or redundant information.

\textbf{Anchoring Knowledge via an Evolving Index.} Unlike prior methods, \method{} introduces an evolving in-context index $I$ as a structured mechanism to anchor key information in the retrieved knowledge. This index supports reasoning, next-query generation, and final answer prediction in a unified manner. We denote the complete output of the LLM at the $(t-1)$-th step as $o_{t-1}=\{I_{t-1}, R_{t-1}, q_{t-1}\}$, which includes the index, the reasoning process, and the sub-query, respectively.

At step $t$, the LLM generates the updated index $I_t$ and the reasoning process $R_t$ based on the newly retrieved documents $D_t$:
\begin{equation}\label{eq:index_update}
\{I_t, R_t\} = \mathcal{LLM}(q_0, D_t, o_{t-1}),
\end{equation}
where $I_t$ anchors the newly retrieved knowledge $D_t$, and $R_t$ captures both the reasoning process and a knowledge sufficiency judgment. This judgment, classified as either \emph{sufficient} or \emph{insufficient}, determines whether the iterative retrieval should continue. If the judgment is \emph{insufficient}, the model generates the next query $q_t$ conditioned on $I_t$, $D_t$, and $o_{t-1}$:
\begin{equation}
q_t = \mathcal{LLM}(q_0, D_t, o_{t-1}, I_t, R_t).
\end{equation}

Once the iterative retrieval process concludes, all retrieved documents are aggregated to form the complete document set $\mathcal{D}$ (Eq.~\ref{eq:all_document}). The LLM then produces the final answer $y$ using both $\mathcal{D}$ and the final evolving index $I_k$:
\begin{equation}
    y = \mathcal{LLM}(q_0, \mathcal{D}, I_k).
\end{equation}
By progressively refining the index $I$ at each iteration (see Sec.~\ref{section:index} for details), \method{} effectively preserves salient information as a knowledge anchor. Grounding reasoning on this in-context index enables the model to better organize and interpret retrieved knowledge, improving reasoning accuracy and generating more reliable answers.

\subsection{Stepwise Index Evolution for Context-Aware Knowledge Indexing}\label{section:index}
In this subsection, we first introduce the definition and representation of the in-context index $I$ (Sec.~\ref{section:preliminary}), and then present our stepwise index update mechanism.

\textbf{Index Representation.}
During index generation, the LLM is tasked with extracting keywords and relations among them.
Specifically, the index is defined as $I = (\mathcal{K}, \mathcal{R})$, where $\mathcal{K} = \{k_1, \dots, k_n\}$ denotes the set of keywords, and $\mathcal{R} = \{r_1, \dots, r_m\}$ denotes the set of relations extracted from documents, represented as RDF triples~\cite{klyne2004resource}.

Following prior work~\cite{li2023sequence, jain2024structsum}, we linearize the index $I$ by verbalizing both its keywords and triples:
\begin{equation}
\begin{aligned}
& \text{Keywords: } X(k_1), X(k_2), \dots, X(k_n); \\
& \text{Relationships: } X(r_1), X(r_2), \dots, X(r_m),
\end{aligned}
\end{equation}
where $X(k_i)$ denotes a keyword verbalization function that encodes the keyword name, and $X(r_j)$ denotes the verbalized RDF triple $r_j$, consisting of a head keyword, a relation, and a tail keyword.

\paragraph{Stepwise Index Evolution.}
During iterative retrieval, \method{} adopts a stepwise index evolution mechanism to maintain the index $I$, which functions as a dynamic knowledge index for both model reasoning and subsequent query generation.

As formalized in Eq.~\ref{eq:index_update}, the index at step $t$ is updated based on the previous index $I_{t-1}$, the reasoning trace $R_{t-1}$, and the generated query $q_{t-1}$:
\begin{equation}
I_t = \mathcal{LLM}(q_0, D_t, \{I_{t-1}, R_{t-1}, q_{t-1}\}).
\end{equation}
Specifically, when transitioning from $I_{t-1}$ to $I_t$, the index is incrementally updated by incorporating the retrieved documents $D_t$ at step $t$, resulting in a stepwise evolutionary process:
\begin{equation}
\Delta I_{t-1 \rightarrow t} = \text{Index}(D_t, \{R_{t-1}, q_{t-1}\}).
\end{equation}
where $\text{Index}(\cdot)$ denotes key information indexing operation conditioned on the prior reasoning trace $R_{t-1}$ and the previously issued query $q_{t-1}$. These two signals jointly guide the extraction of salient, query-relevant information from $D_t$ for index updating. As a result, the index progressively integrates query-aware evidence across retrieval steps, allowing key information to be incrementally identified, highlighted, and anchored within the evolving index structure.

\section{Experimental Methodology}
In this section, we describe the datasets, baselines, evaluation metrics, and implementation details in our experiments. More experimental details are shown in Appendix~\ref{appendix:add_exp}.

\textbf{Datasets.} Following previous work~\cite{shao2023enhancing}, we use four multi-hop question answering datasets to evaluate the effectiveness of our method, including MuSiQue~\cite{trivedi2022musique}, HotpotQA~\cite{yang2018hotpotqa}, 2WikiMultihopQA (2WikiMQA)~\cite{ho2020constructing}, and Bamboogle~\cite{press2022measuring}. 

\textbf{Baselines.} In our experiments, we compare \method{} with the Vanilla RAG model~\cite{ram2023context}, three GraphRAG baselines (\texttt{GraphRAG}~\cite{edge2024local}, \texttt{LightRAG}~\cite{guo2024lightrag}, and \texttt{HippoRAG2}~\cite{DBLP:conf/icml/GutierrezSQZ025}), and four iterative retrieval methods (\texttt{IRCoT}~\cite{trivedi2023interleaving}, \texttt{Iter-RetGen}~\cite{shao2023enhancing}, \texttt{DeepNote}~\cite{wang2025deepnotenotecentricdeepretrievalaugmented}, and \texttt{Search-R1}~\cite{jin2025search}). More description about baseline details, please refer to Appendix~\ref{appendix:add_exp}.



\textbf{Evaluation Metrics.} Following~\citet{wang2025deepnotenotecentricdeepretrievalaugmented}, we utilize Token-Level F1 (F1) and Exact Match (EM) as evaluation metrics. F1 score captures the trade-off between content coverage and redundancy, allowing for partial matches between the predicted and reference answers. Exact Match (EM) focuses on whether the model’s output exactly and completely matches the reference answer, providing a stricter measure of precision.

\textbf{Implementation Details.} We implemented \method{} using Qwen2.5-7B-Instruct~\cite{qwen2} and Llama3.1-8B-Instruct~\cite{dubey2024llama} as the backbone models. During retrieval, bge-large-en-v1.5~\cite{xiao2024c} is used to retrieve query-related documents, and the top 5 retrieved documents are fed into the LLM at each retrieval step. The maximum number of retrieval steps was limited to 4. 
All instruction templates used in \method{} are shown in Appendix~\ref{appendix:add_exp}.

\begin{table*}[ht]
\centering
\resizebox{\linewidth}{!}{
\begin{tabular}{lcccccccccc}
\toprule 
\multirow{2}{*}{\textbf{Method}} & \multicolumn{2}{c}{\textbf{MuSiQue}} & \multicolumn{2}{c}{\textbf{HotpotQA}} & \multicolumn{2}{c}{\textbf{2WikiMQA}} &\multicolumn{2}{c}{\textbf{Bamboogle}} &\multicolumn{2}{c}{\textbf{Average}}\\ 
&\textbf{F1} & \textbf{EM} & \textbf{F1}  & \textbf{EM}  & \textbf{F1} & \textbf{EM}  & \textbf{F1} & \textbf{EM} & \textbf{F1} & \textbf{EM} \\ 
\midrule
\multicolumn{11}{l}{\textit{Qwen2.5-7B-Instruct}}\\
\hdashline
Vanilla RAG~\shortcite{ram2023context} &{18.05} &{9.80} &{50.93} &{40.40} &{38.54} &{33.60} &{21.62} &{13.60} &{32.29} &{24.35} \\
GraphRAG~\shortcite{edge2024local} &7.70 &4.08 &41.27 &38.00 &31.83 &28.00 &22.14 &20.00 &25.83 &22.61\\
LightRAG~\shortcite{guo2024lightrag} &15.00 & 4.08 & 45.73 & 36.00 & 34.77 &30.00 &19.84 &14.00 &28.90 &21.11 \\
HippoRAG2~\shortcite{DBLP:conf/icml/GutierrezSQZ025} &12.99 &8.16 &41.27 &36.00 &25.02 &24.00 &18.27 &16.00 &24.44 &21.11 \\
IRCoT~\shortcite{trivedi2023interleaving} &{17.91} &{8.20} &{49.55} &{36.60} &{43.47} &{32.80} &{21.15} &{12.80} &{33.02} &{22.60} \\
Iter-RetGen~\shortcite{shao2023enhancing} &{21.36} &{13.00} &{58.81} &{47.40} &{43.15} &{37.60} &{24.82} &{16.80} &{37.04} &{28.70}  \\
Search-R1~\shortcite{jin2025search} &{24.95} &{17.60} &{54.78} &{43.40} &{49.55} &{40.60} &{25.13} &{20.00} &{38.60} &{30.40}  \\
DeepNote~\shortcite{wang2025deepnotenotecentricdeepretrievalaugmented} &{26.67} &{15.80} &{59.97} &{48.60} &{53.72} &{43.40} &{32.66} &{24.00} &{43.26} &{32.95} \\
\method{} &\textbf{31.17} &\textbf{21.20} &\textbf{62.64} &\textbf{50.80} &\textbf{60.99} &\textbf{52.21} &\textbf{36.28} &\textbf{27.20} &\textbf{47.77} &\textbf{37.85}\\

\midrule
\multicolumn{11}{l}{\textit{Llama3.1-8B-Instruct}}\\
\hdashline
Vanilla RAG~\shortcite{ram2023context} &{17.43} &{9.40} &{52.94} &{42.20} &{37.64} &{32.60} &{24.40} &{16.00} &{33.18} &{25.05} \\
GraphRAG~\shortcite{edge2024local} &11.84 &10.20 &35.80 &28.00 &23.89 &20.00 &22.27 &18.00 &23.51 &19.10\\
LightRAG~\shortcite{guo2024lightrag} &9.93 &2.04 &55.49 &44.00 &36.13 &34.00 &22.00 &16.00 &30.99 &24.12 \\
HippoRAG2~\shortcite{DBLP:conf/icml/GutierrezSQZ025} &19.76 &12.24 &36.17 &24.00 &24.87 &22.00 &19.60 &18.00 &25.13 &19.10\\
IRCoT~\shortcite{trivedi2023interleaving} &{19.91} &{12.40} &{50.41} &{39.60} &{49.67} &{39.20} &{32.46} &{23.20} &{38.11} &{28.60}\\
Iter-RetGen~\shortcite{shao2023enhancing} &{18.15} &{11.20} &{55.97} &{43.80} &{34.32} &{26.60} &{26.09} &{20.00} &{33.63} &{25.40}\\
DeepNote~\shortcite{wang2025deepnotenotecentricdeepretrievalaugmented} &{26.81} &{16.60} &{59.51} &{47.80} &{56.65} &{45.80} &{37.46} &{27.20}&{45.11} &{34.35} \\
\method{} &\textbf{34.05} &\textbf{21.56} &\textbf{68.34} &\textbf{54.91} &\textbf{63.45} &\textbf{56.07} &\textbf{41.35} &\textbf{30.91} &\textbf{51.80} &\textbf{40.86}\\

\bottomrule
\end{tabular}}
\caption{\label{table/overall} Overall Performance. We adopt Qwen2.5-7B-Instruct and Llama3.1-8B-Instruct as backbone models for evaluation. Since Search-R1 does not provide an open-source implementation based on Llama3.1-8B-Instruct, the corresponding results are therefore not reported.}
\end{table*}


\section{Evaluation Result}
In this section, we first evaluate the performance of various iterative RAG reasoning methods, followed by an ablation study to investigate the contribution of individual modules in \method{}. 
We then show the effectiveness of index in \method{} during answer generation and iterative retrieval.
In addition, we show the attention visualization in \method{} during iterative retrieval in Appendix~\ref{appendix:attention}. Finally, we show a case study in Appendix~\ref{appendix:case_study}.

\subsection{Overall Performance}
As shown in Table~\ref{table/overall}, we report the performance of \method{} along with several representative RAG baseline models. The evaluation results in Table~\ref{table/overall} demonstrate the consistent effectiveness of \method{} across different datasets, outperforming Vanilla RAG and GraphRAG-based methods by approximately a 20\% improvement.


Compared with deep retrieval-based methods such as IRCoT, Iter-RetGen, and Search-R1, \method{} achieves approximately 10\% improvements. This highlights the critical role of the constructed index in facilitating effective understanding and utilization of retrieved knowledge across multiple retrieval iterations. Benefiting from the index evolution process, \method{} is able to utilize retrieved knowledge and provides a mechanism for anchoring essential information, thereby mitigating the negative impact of noisy retrieval results. Among all baseline models, the graph-based methods exhibit a slight performance drop compared to vanilla RAG models, whereas iterative retrieval methods achieve significantly better results. This suggests that representing knowledge as graphs may omit essential information needed to assist LLMs in answering questions. In contrast, \method{} shifts the focus from knowledge representations to in-context knowledge anchoring through an evolving index, which leads to more substantial performance gains on these QA tasks. Moreover, we explored other indexing formats and reported the iterative RAG performance and their ablation studies in Appendix~\ref{appendix:other_format}.
Further, we evaluate Vanilla RAG under different retrieval depths in Appendix~\ref{appendix:vanilla}.

\begin{table*}[ht]
\centering
\small
\resizebox{\linewidth}{!}{
\begin{tabular}{lcccccccccc}
\toprule 
\multirow{2}{*}{\textbf{Method}} & \multicolumn{2}{c}{\textbf{MuSiQue}} & \multicolumn{2}{c}{\textbf{HotpotQA}} & \multicolumn{2}{c}{\textbf{2WikiMQA}} &\multicolumn{2}{c}{\textbf{Bamboogle}} &\multicolumn{2}{c}{\textbf{Average}}\\ 
  & \textbf{F1}  & \textbf{EM}  & \textbf{F1}  & \textbf{EM}  & \textbf{F1} & \textbf{EM}  & \textbf{F1} & \textbf{EM} & \textbf{F1} & \textbf{EM} \\ 
\midrule
\rowcolor{gray!8} \multicolumn{11}{l}{\textit{Qwen2.5-7B-Instruct}}\\
\hdashline
Vanilla RAG &{18.05} &{9.80} &{50.93} &{40.40} &{38.54} &{33.60} &{21.62} &{13.60} &{32.29} &{24.35} \\
\hdashline
\method{} (QA w/ Docs) &{25.16} &{14.00} &{62.30} &{50.20} &{51.97} &{43.57} &{28.74} &{20.00} &{42.04} &{31.94} \\
\method{} (QA w/ Index) &{23.18} &{15.00} &{59.39} &{48.20} &{58.25} &{49.40} &{35.74} &\textbf{28.00} &{44.14} &{35.15} \\
\method{} &\textbf{31.17} &\textbf{21.20} &\textbf{62.64} &\textbf{50.80} &\textbf{60.99} &\textbf{52.21} &\textbf{36.28} &{27.20} &\textbf{47.77} &\textbf{37.85}\\
w/o Index &{22.99} &{12.63} &{55.05} &{43.60} &{50.64} &{42.80} &{26.66} &{18.40} &{38.84} &{29.36} \\
w/o Evolving &{22.11} &{11.40} &{58.07} &{45.67} &{41.87} &{31.80} &{22.98} &{14.40} &{36.26} &{25.82}\\
\midrule
\rowcolor{gray!8} \multicolumn{11}{l}{\textit{Llama3.1-8B-Instruct}}\\
\hdashline
Vanilla RAG &{17.43} &{9.40} &{52.94} &{42.20} &{37.64} &{32.60} &{24.40} &{16.00} &{33.18} &{25.05}\\
\hdashline
\method{} (QA w/ Docs) &{28.17} &{17.96} &{64.25} &{51.64} &{52.44} &{44.77} &{32.12} &{25.45} &{44.25} &{34.96} \\
\method{} (QA w/ Index) &{30.02} &{16.17} &{63.03} &{49.82} &{60.77} &{53.56} &{41.03} &{27.27} &{48.71} &{36.71} \\
\method{} &\textbf{34.05} &\textbf{21.56} &\textbf{68.34} &\textbf{54.91} &\textbf{63.45} &\textbf{56.07} &\textbf{41.35} &\textbf{30.91} &\textbf{51.80} &\textbf{40.86}\\
w/o Index &{21.46} &{13.48} &{63.03} &{49.64} &{46.82} &{38.59} &{33.03} &{26.92} &{41.09} &{32.16} \\
w/o Evolving &{21.53} &{13.24} &{61.15} &{49.40} &{45.36} &{34.35} &{32.57} &{27.50} &{40.15} &{31.12} \\
\bottomrule
\end{tabular}}
\caption{\label{table/ablation} Ablation Study.}
\end{table*}
\subsection{Ablation Study}
To better understand the contribution of each core component in \method{}, we conduct ablation studies as shown in Table~\ref{table/ablation}. 

Four ablation variants are compared. First, we evaluate \method{} (QA w/ Docs) and \method{} (QA w/ Index) to assess the knowledge utilization capability by prompting LLMs to answer questions based solely on retrieved documents and the constructed index, respectively. Then, we conduct \method{} w/o Index to analyze the role of the index component. Finally, we conduct \method{} w/o Evolving to demonstrate the importance of index evolution during iterative retrieval, in which we construct the index based on all retrieved knowledge using \method{} w/o Index and use the index and retrieved knowledge to answer the question. 

When question answering relies only on the constructed Index (\method{} (QA w/ Index)), the RAG performance drops by approximately 3\%. This indicates that the index alone does not provide more effective knowledge for directly answering queries; instead, it serves as a form of knowledge anchoring that helps LLMs better interpret the retrieved information. We further analyze the effects of removing the evolution and index modules. When the index is removed from both retrieval and QA stages of \method{}, rather than only from the QA stage (\method{} (QA w/ Docs)), the RAG performance decreases more substantially, by over 8\% (about 5\% more than the drop observed in \method{} (QA w/ Docs)). This result indicates that the index structure further facilitates iterative retrieval by enabling the generation of more effective subqueries, which in turn retrieve more relevant knowledge. In addition, the significant performance drop observed in the w/o Evolving setting, averaging a decrease of over 10\%, which highlights that the value of indexing lies not just in summarization, but in its dynamic accumulation of knowledge. Without iterative evolution, the model struggles to bridge the logical gaps between retrieved knowledge in different stages, leading to a fragmented understanding of the multi-hop reasoning tasks.

\subsection{Effectiveness of Index in \method{} during Answer Generation}
In this subsection, we show the effectiveness of the index during answer generation by testing its performance stability when shifting the placement of critical evidence, and by demonstrating its overall robustness against random document shuffling and reordering. We conduct experiments on the full HotpotQA dataset using the Qwen2.5-7B-Instruct model.

\begin{figure}[t]
    \centering
    \subfigure[Hit Rate of Golden Documents.] { \label{fig:passage_hit_rate} 
    \includegraphics[width=0.44\linewidth]{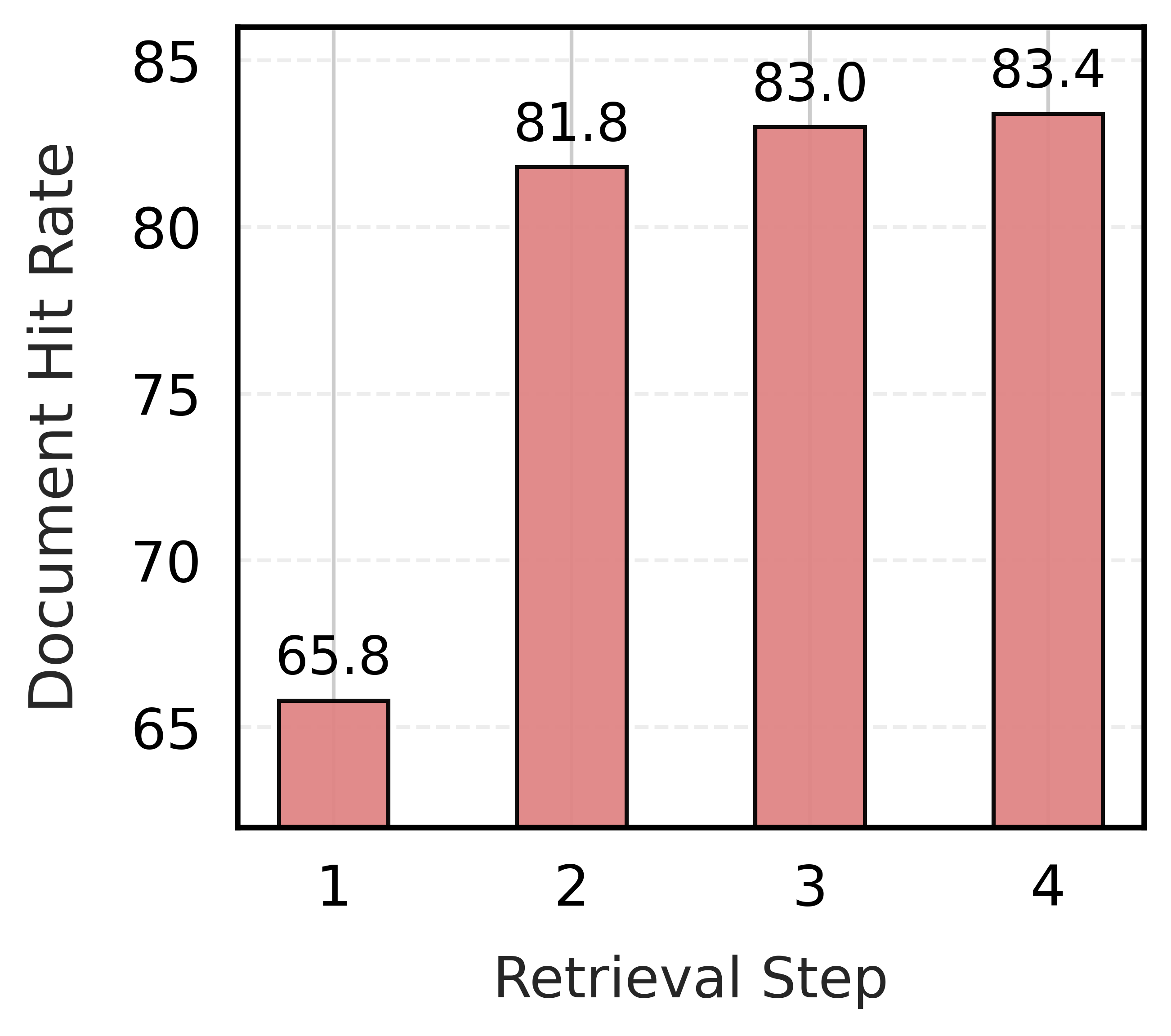}}
    \subfigure[RAG Performance with Postion Disturbance.] { \label{fig:position_analysis} 
    \includegraphics[width=0.44\linewidth]{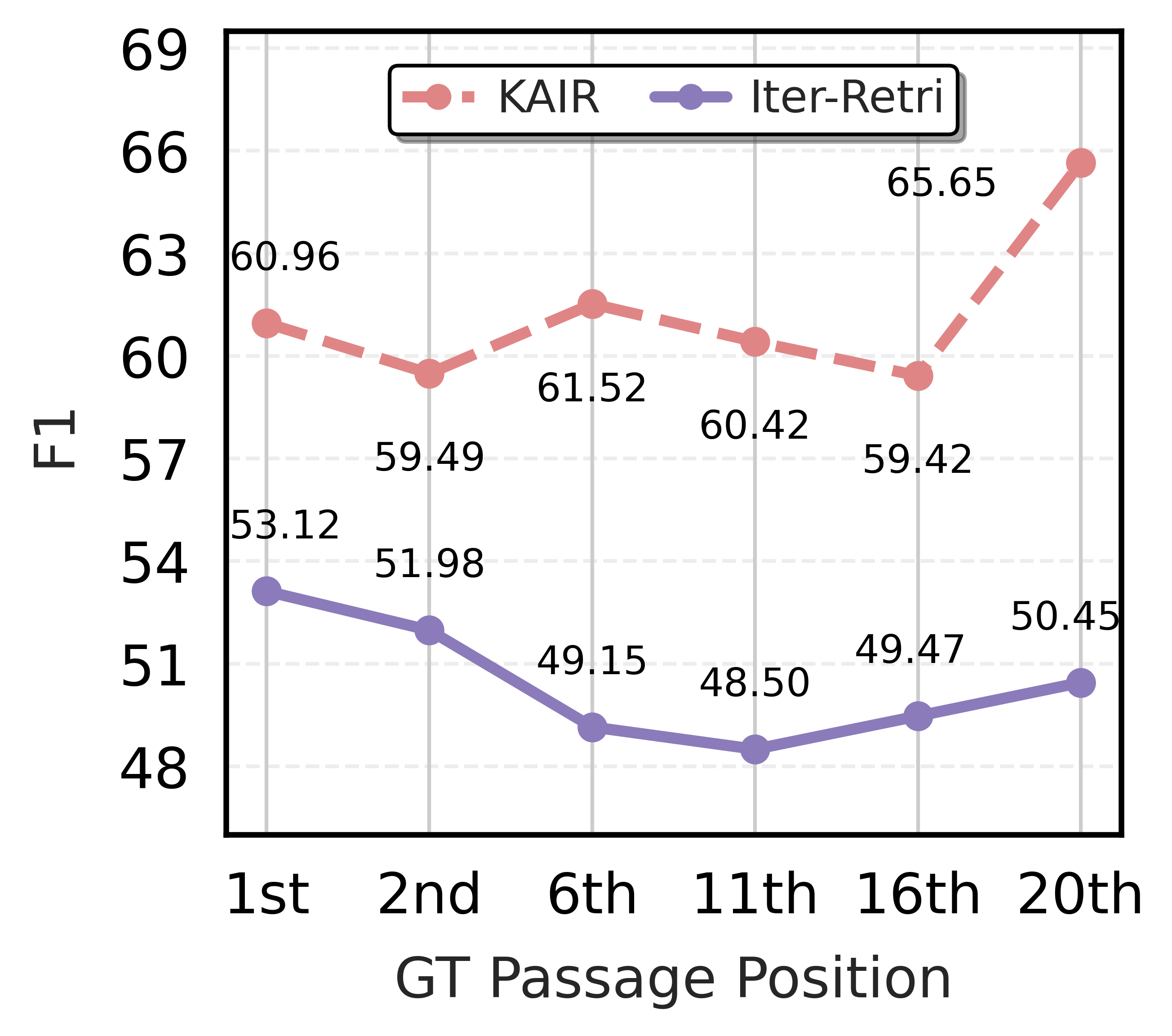}}
    \caption{Mitigate Iterative Retrieval Noise. Iter-Retri denotes standard iterative retrieval.}
    \label{fig:position_hit_rate}
\end{figure}

\textbf{Mitigate Iterative Retrieval Noise.}
In Figure~\ref{fig:position_hit_rate}, we analyze the hit rate of the golden document and the RAG performance with position disturbance to show the effective ability of \method{} in mitigating iterative retrieval noise.

First, Figure~\ref{fig:passage_hit_rate} calculates the cumulative document hit rate of the golden evidence across retrieval iterations. While the hit rate increases progressively with each retrieval step, confirming that sequential retrieval successfully expands knowledge coverage, it quickly plateaus after the second round. This trend indicates that deeper retrieval yields diminishing returns in capturing new target documents, while inevitably introducing noise information into the context. As depicted in the Figure~\ref{fig:position_analysis}, we explicitly manipulate the placement of documents with the ground truth answer in the evidence list. Iter-Retri exhibits a pronounced vulnerability to this displacement, suffering from a severe ``lost in the middle'' phenomenon~\cite{liu2023lost} that drastically impairs in-context knowledge utilization. In contrast, \method{} maintains a consistently stable and superior performance regardless of the absolute position of ground truth documents. This clearly demonstrates that the dynamic index successfully acts as a global cognitive anchor, immunizing the model against the arbitrary positioning of critical facts and redundant information. 




\begin{figure}[t]
    \centering
    \subfigure[Retrieval Shuffling.] { \label{fig:robustness} 
    \includegraphics[width=0.46\linewidth]{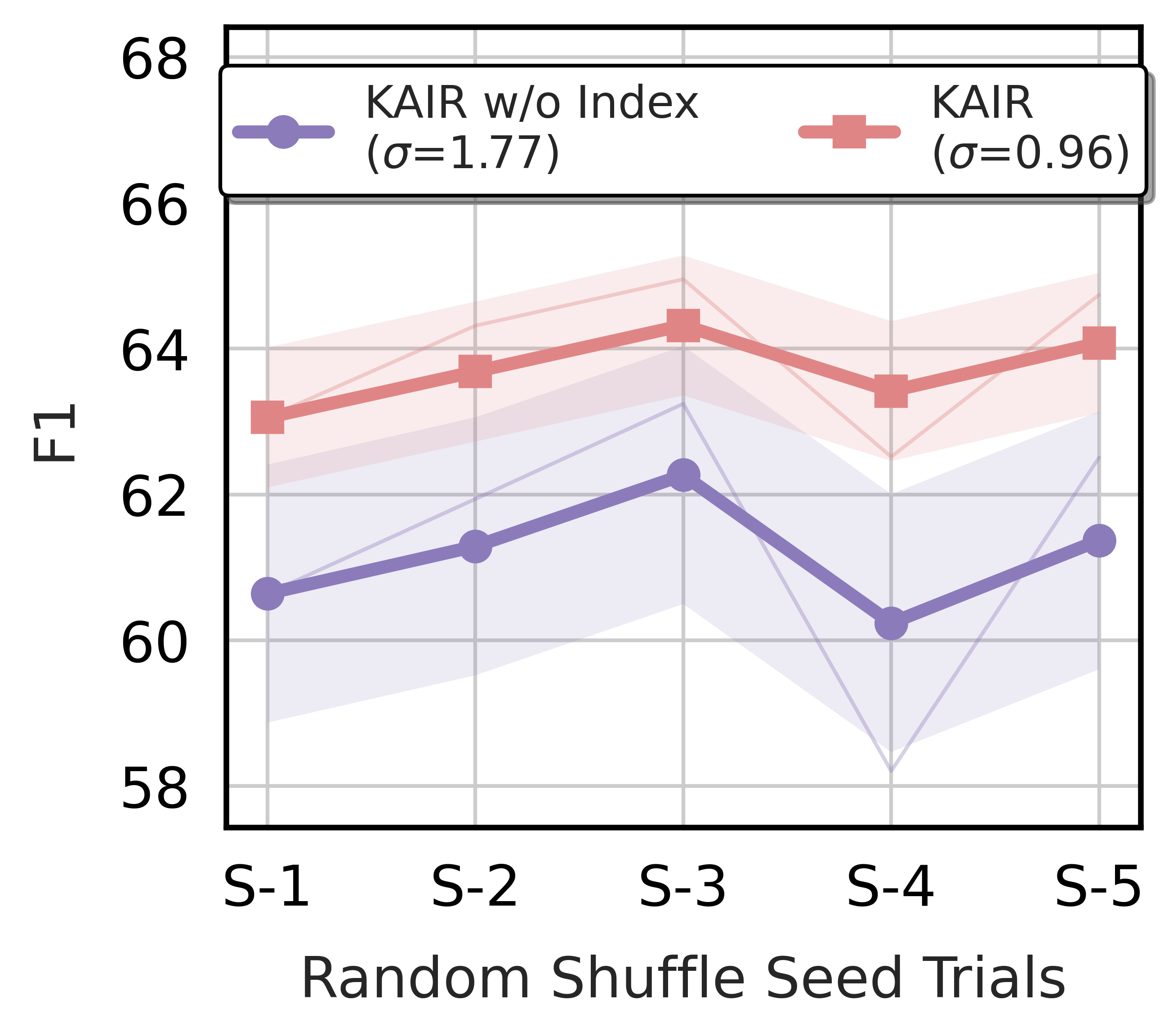}}
    \subfigure[Re-ranking and Retrieval Reordering.] { \label{fig:arrangement_strategies} 
    \includegraphics[width=0.46\linewidth]{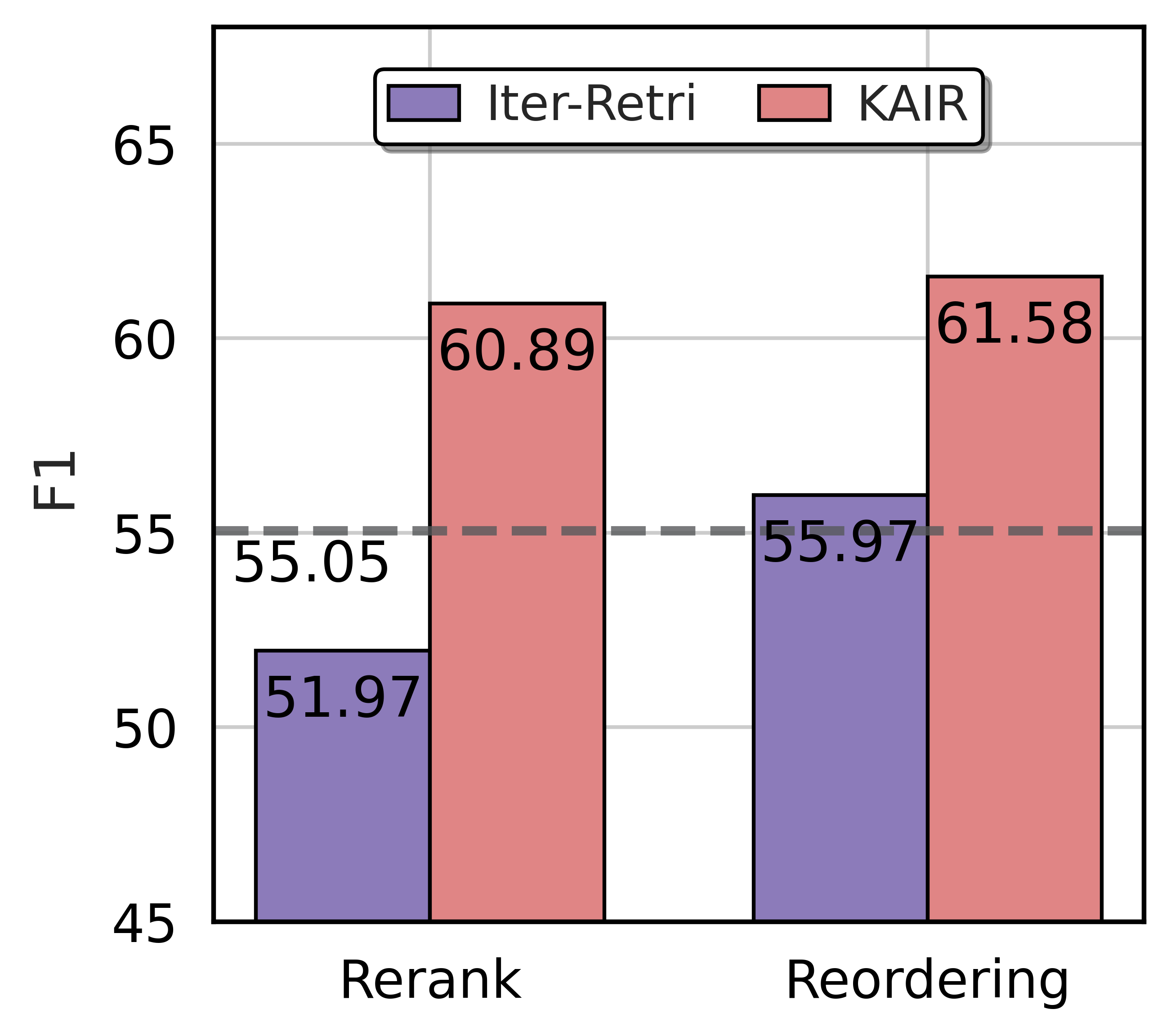}}
    \caption{RAG Robustness with Different Retrieval Ordering Strategies. Iter-Retri denotes standard iterative retrieval. Grey line denotes w/o reordering method.}
    \label{fig:robustness_volatility}
\end{figure}

\textbf{RAG Robustness of \method{}.} 
Figure~\ref{fig:robustness_volatility} illustrates the robustness of \method{} through different retrieval ordering strategies.

First, to evaluate the robustness of \method{} in anchoring key information within retrieved documents, we randomly shuffle retrieved passages five times (the tails are from S-1 to S-5) and show the performance variance of \method{} w/o Index and \method{}. As illustrated in Figure~\ref{fig:robustness}, Iter-Retri is highly sensitive to the input context's order. Its RAG performance fluctuates significantly across different trials, accompanied by a wide shaded confidence band. This indicates that without knowledge anchoring, the RAG performance is heavily dictated by the arbitrary positioning of relevant information. In contrast, \method{} not only consistently achieves a higher overall performance but also exhibits a much flatter performance curve with a notably narrower shaded band. 
We also compare the standard deviation ($\sigma$) of the F1 scores across the trials. \method{} drastically reduces the performance volatility from $\sigma = 1.77$ to $\sigma = 0.96$. 
By organizing and tracking the accumulated knowledge, \method{} successfully anchors the key information within retrieved knowledge, rendering the robust performance regardless of how the retrieved documents are sequenced.

Furthermore, we explore whether different retrieval document reordering strategies can affect the RAG performance. Figure~\ref{fig:arrangement_strategies} illustrates the performance under semantic reranking and retrieval reordering~\cite{DBLP:conf/iclr/JinY0A25}. Strikingly, without using an index, the model suffers a catastrophic degradation under different reordering strategies: performance plummets to 51.97 with reranking, whereas retrieval reordering offers a marginal benefit, reaching 55.97. While the model uses the index to anchor the silent information, its RAG performance improves significantly regardless of the different reordering strategies.

\subsection{Effectiveness of Index in \method{} during Iterative Retrieval}\label{section:characteristics_index}
This subsection analyzes the effectiveness of the index constructed by \method{} in the iterative retrieval process. We conduct experiments on the full HotpotQA dataset using the Qwen2.5-7B-Instruct model to show the role of the index in knowledge anchoring. Then, we explore how the constructed index assists LLMs in anchoring knowledge through the attention mechanism. Instances from HotpotQA that require the maximum number of retrieval steps (4 steps) are selected to illustrate the knowledge anchoring mechanism.

\begin{figure}[t]
    \centering 
     \subfigure[Index Changes with Different Retrieval Steps.] { \label{fig:keywordgrowth_hitrate} 
    \includegraphics[width=0.48\linewidth]{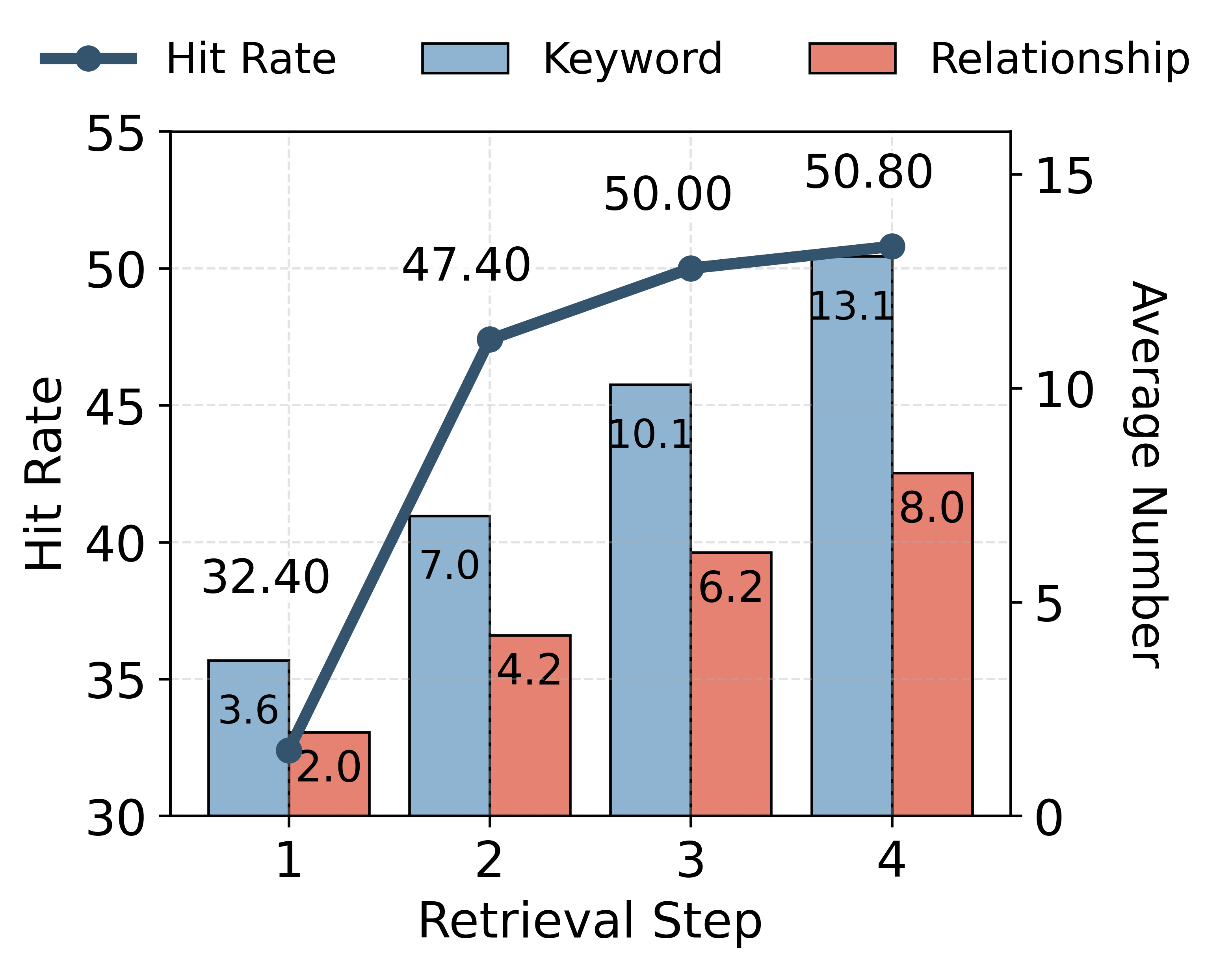}}
    \subfigure[Effectiveness of Constructed Indices.] { \label{fig:performance} 
    \includegraphics[width=0.45\linewidth]{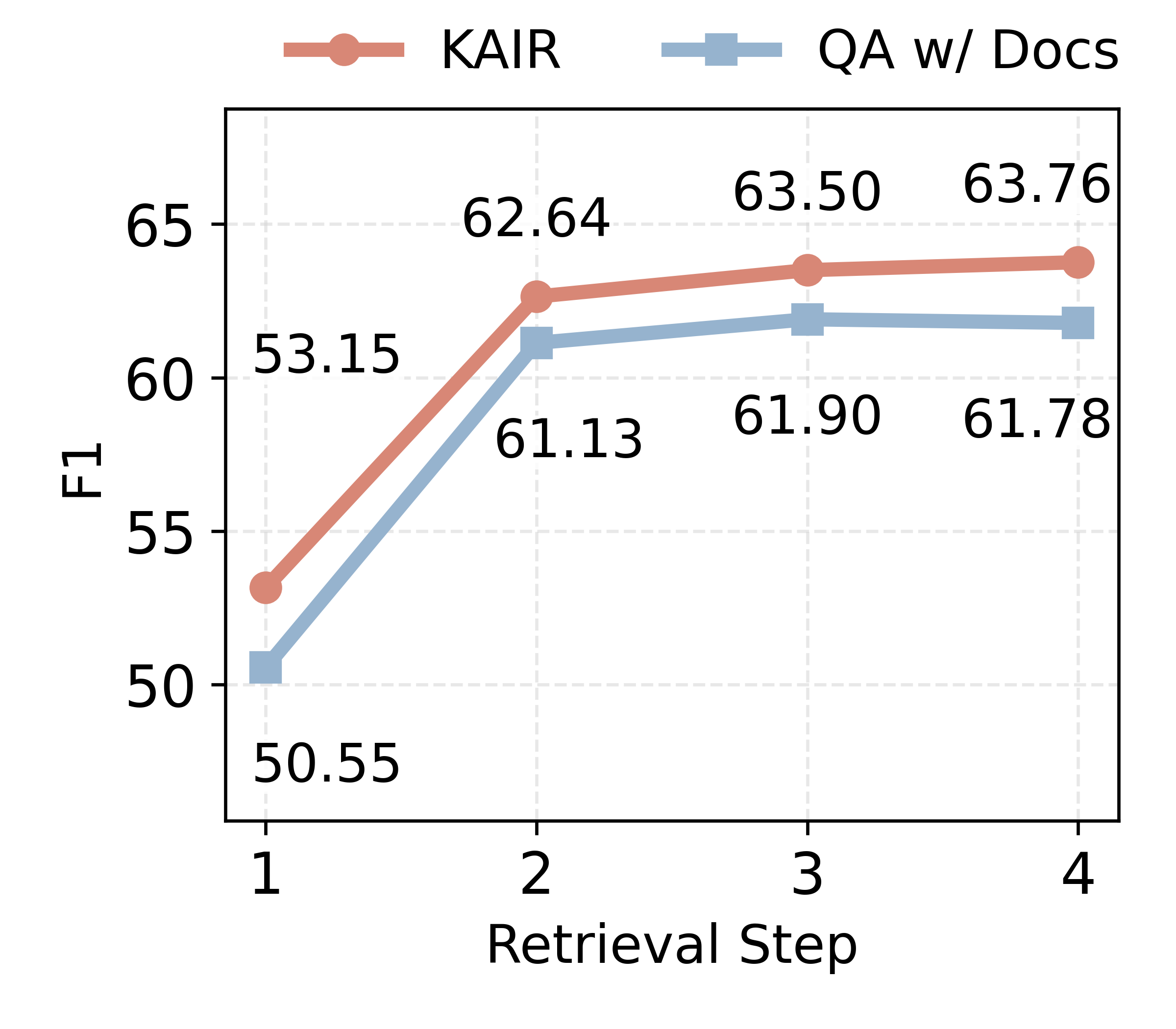}}
    \caption{Knowledge Anchoring Effectiveness across Different Retrieval Steps.}
    \label{fig:index_performance}
\end{figure}
\textbf{Index Effectiveness in Knowledge Anchoring.}
As shown in Figure~\ref{fig:index_performance}, we illustrate the knowledge anchoring effectiveness of the constructed index at each retrieval step by analysing index evolution and reporting the performance of RAG models as additional retrieved documents are incorporated. 

First, Figure~\ref{fig:keywordgrowth_hitrate} presents the index changes across different retrieval steps. Specifically, as the iterations proceed, the number of keywords and relationships steadily increases, indicating that the index incrementally extracts key information from retrieved documents for indexing purposes. Notably, this growth continues to improve after the third retrieval step, while the hit rate of the ground-truth answer reaches a plateau. This observation suggests that the indexing process does not merely incorporate query-related keywords and relationships, but instead dynamically adjusts the index structure to enable more effective knowledge indexing.
Then, we evaluate the RAG performance at different retrieval steps using the F1 score in Figure~\ref{fig:performance}. Compared with the \method{} (QA w/ Docs) model, \method{} consistently outperforms it across retrieval steps. This observation indicates that the benefits of the constructed index are amplified with a deeper retrieval step, enabling effective utilization of accumulated knowledge.

\begin{figure}[t]
    \centering
    \subfigure[QA Perplexity.] { \label{fig:ppl} 
    \includegraphics[width=0.48\linewidth]{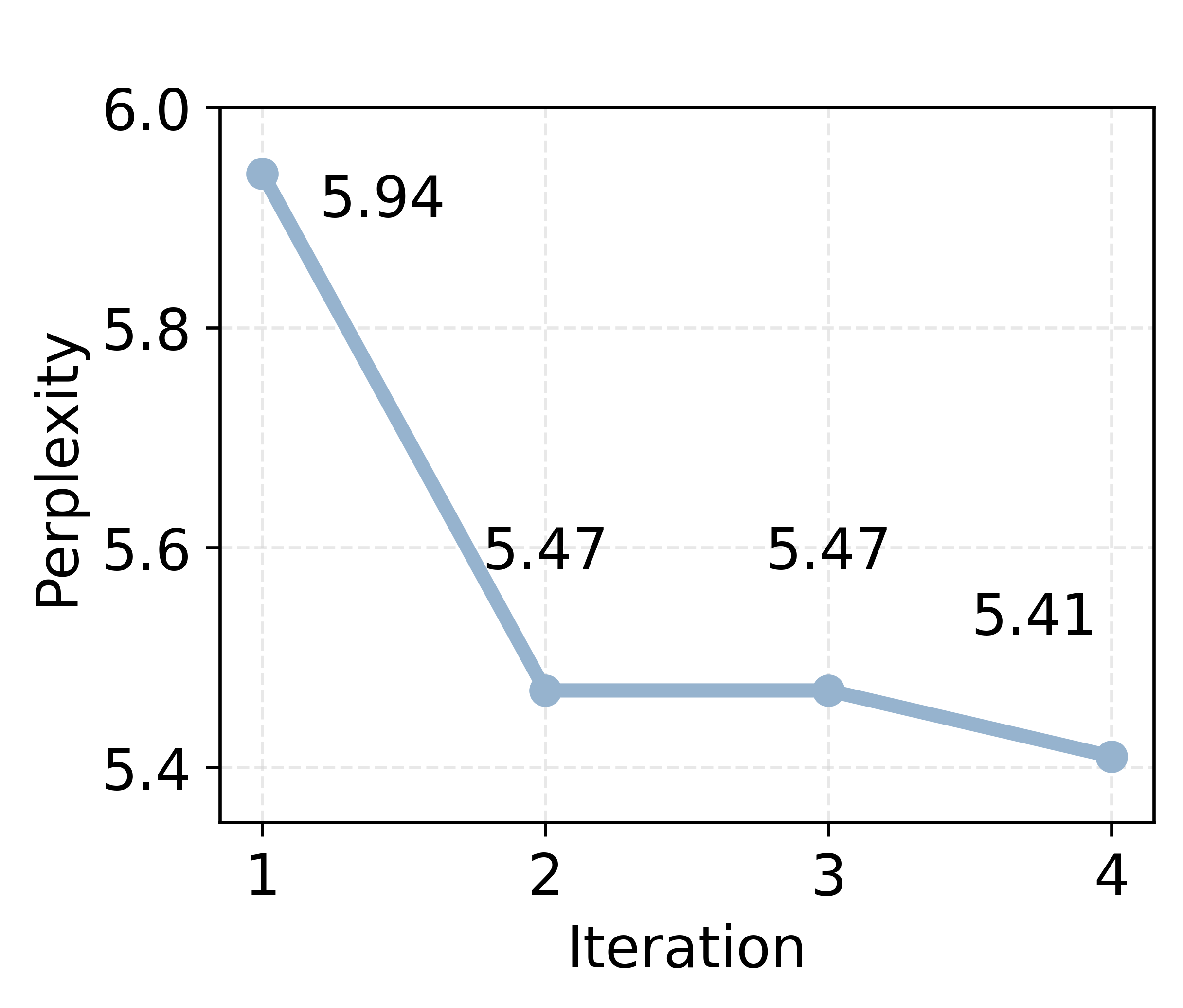}}
    \subfigure[Attention Weight of New Index Keywords.] { \label{fig:stepwise_lighting} 
    \includegraphics[width=0.48\linewidth]{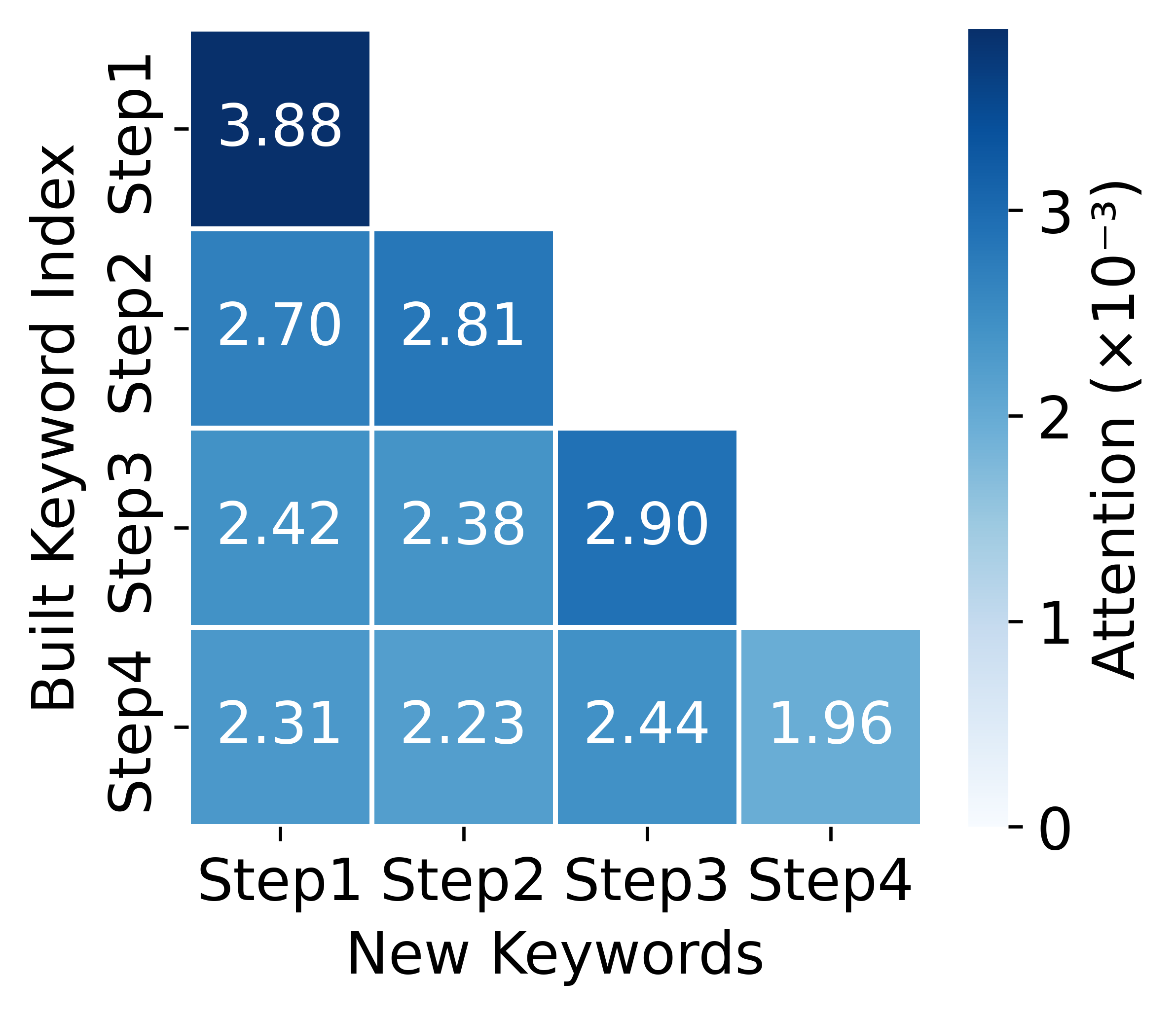}}
    \caption{Attention Distribution on Indexed Keywords.}
    \label{fig:mixed}
\end{figure}

\textbf{Knowledge Anchoring via In-Context Indexing.} Figure~\ref{fig:mixed} shows the QA perplexity under incremental passage accumulation and illustrates the indexing mechanism, which computes attention weights over anchored keywords mentioned in the retrieved documents.

As shown in Figure~\ref{fig:ppl}, the perplexity of ground-truth answer generation steadily decreases, suggesting that the evolving index indeed helps LLMs leverage the retrieved documents more effectively for question answering. Moreover, as shown in Figure~\ref{fig:stepwise_lighting}, we further examine the attention weights of the index-anchored keywords that are newly extracted from documents at different retrieval steps. The peak intensity along the diagonal indicates that, at each retrieval step, the newly index-anchored keywords receive substantially higher attention once the index constructed at the corresponding retrieval iteration is incorporated. This observation suggests that the step-by-step indexing effectively modulates the attention distribution of LLMs to facilitate more accurate knowledge interpretation and utilization.

\section{Conclusion}
This paper introduces \method{}, a novel knowledge anchoring approach that maintains an evolutive index during iterative retrieval. By dynamically updating the index at each retrieval step, the proposed index enables LLMs to anchor a broader set of related information for knowledge utilization, enabling more reliable retrieval and robust answer generation.


\section*{Limitations}
Although \method{} demonstrates effectiveness in knowledge anchoring for RAG systems, the quality of the constructed index remains constrained by the capabilities of the underlying LLM. Specifically, the performance of the proposed index construction and utilization depends on the model's ability to accurately extract, structure, and organize relevant information from retrieved documents. In addition, future work could explore and identify the optimal index formats for different types of questions.


\bibliography{reference_norm}

@article{FlashRAG,
  author       = {Jiajie Jin and
                  Yutao Zhu and
                  Xinyu Yang and
                  Chenghao Zhang and
                  Zhicheng Dou},
  title        = {FlashRAG: {A} Modular Toolkit for Efficient Retrieval-Augmented Generation
                  Research},
  journal      = {Arxiv preprint},
  volume       = {abs/2405.13576},
  year         = {2024},
  url          = {https://arxiv.org/abs/2405.13576},
}

@inproceedings{xiao2024c,
 author = {Xiao, Shitao and Liu, Zheng and Zhang, Peitian and Muennighoff, Niklas and Lian, Defu and Nie, Jian-Yun},
 booktitle = {Proceedings of the 47th international ACM SIGIR conference on research and development in information retrieval},
 pages = {641--649},
 title = {C-pack: Packed resources for general chinese embeddings},
 year = {2024},
 url = {https://dl.acm.org/doi/abs/10.1145/3626772.3657878}
}

@article{qwen2,
 author = {An Yang and Baosong Yang and Binyuan Hui and Bo Zheng and Bowen Yu and Chang Zhou and Chengpeng Li and Chengyuan Li and Dayiheng Liu and Fei Huang and Guanting Dong and Haoran Wei and Huan Lin and Jialong Tang and Jialin Wang and Jian Yang and Jianhong Tu and Jianwei Zhang and Jianxin Ma and Jin Xu and Jingren Zhou and Jinze Bai and Jinzheng He and Junyang Lin and Kai Dang and Keming Lu and Keqin Chen and Kexin Yang and Mei Li and Mingfeng Xue and Na Ni and Pei Zhang and Peng Wang and Ru Peng and Rui Men and Ruize Gao and Runji Lin and Shijie Wang and Shuai Bai and Sinan Tan and Tianhang Zhu and Tianhao Li and Tianyu Liu and Wenbin Ge and Xiaodong Deng and Xiaohuan Zhou and Xingzhang Ren and Xinyu Zhang and Xipin Wei and Xuancheng Ren and Yang Fan and Yang Yao and Yichang Zhang and Yu Wan and Yunfei Chu and Yuqiong Liu and Zeyu Cui and Zhenru Zhang and Zhihao Fan},
 journal = {ArXiv preprint},
 title = {Qwen2 Technical Report},
 url = {https://arxiv.org/abs/2407.10671},
 year = {2024}
}

@article{dubey2024llama,
 author = {Dubey, Abhimanyu and Jauhri, Abhinav and Pandey, Abhinav and Kadian, Abhishek and Al-Dahle, Ahmad and Letman, Aiesha and Mathur, Akhil and Schelten, Alan and Yang, Amy and Fan, Angela and others},
 journal = {Arxiv preprint},
 pages = {arXiv--2407},
 title = {The llama 3 herd of models},
 year = {2024},
 url = {https://arxiv.org/abs/2407.21783}
}

@article{trivedi2022musique,
 author = {Trivedi, Harsh and Balasubramanian, Niranjan and Khot, Tushar and Sabharwal, Ashish},
 journal = {Transactions of the Association for Computational Linguistics},
 pages = {539--554},
 title = {MuSiQue: Multihop Questions via Single-hop Question Composition},
 year = {2022},
 url = {https://aclanthology.org/2022.tacl-1.31}
}

@inproceedings{yang2018hotpotqa,
 author = {Yang, Zhilin  and
Qi, Peng  and
Zhang, Saizheng  and
Bengio, Yoshua  and
Cohen, William  and
Salakhutdinov, Ruslan  and
Manning, Christopher D.},
 booktitle = {Proceedings of EMNLP},
 pages = {2369--2380},
 title = {{H}otpot{QA}: A Dataset for Diverse, Explainable Multi-hop Question Answering},
 url = {https://aclanthology.org/D18-1259},
 year = {2018}
}

@inproceedings{ho2020constructing,
 author = {Ho, Xanh  and
Duong Nguyen, Anh-Khoa  and
Sugawara, Saku  and
Aizawa, Akiko},
 booktitle = {Proceedings of COLING},
 pages = {6609--6625},
 title = {Constructing A Multi-hop {QA} Dataset for Comprehensive Evaluation of Reasoning Steps},
 url = {https://aclanthology.org/2020.coling-main.580},
 year = {2020}
}

@article{press2022measuring,
 author = {Press, Ofir and Zhang, Muru and Min, Sewon and Schmidt, Ludwig and Smith, Noah A and Lewis, Mike},
 journal = {ArXiv preprint},
 title = {Measuring and narrowing the compositionality gap in language models},
 url = {https://arxiv.org/abs/2210.03350},
 year = {2022}
}

@article{ram2023context,
 author = {Ram, Ori and Levine, Yoav and Dalmedigos, Itay and Muhlgay, Dor and Shashua, Amnon and Leyton-Brown, Kevin and Shoham, Yoav},
 journal = {Transactions of the Association for Computational Linguistics},
 pages = {1316--1331},
 title = {In-context retrieval-augmented language models},
 year = {2023},
 url = {https://aclanthology.org/2023.tacl-1.75/}
}

@article{shi2023replug,
 author = {Shi, Weijia and Min, Sewon and Yasunaga, Michihiro and Seo, Minjoon and James, Rich and Lewis, Mike and Zettlemoyer, Luke and Yih, Wen-tau},
 journal = {ArXiv preprint},
 title = {Replug: Retrieval-augmented black-box language models},
 url = {https://arxiv.org/abs/2301.12652},
 year = {2023}
}

@article{shao2023enhancing,
 author = {Shao, Zhihong and Gong, Yeyun and Shen, Yelong and Huang, Minlie and Duan, Nan and Chen, Weizhu},
 journal = {ArXiv preprint},
 title = {Enhancing retrieval-augmented large language models with iterative retrieval-generation synergy},
 url = {https://arxiv.org/abs/2305.15294},
 year = {2023}
}

@inproceedings{
asai2024selfrag,
title={Self-{RAG}: Learning to Retrieve, Generate, and Critique through Self-Reflection},
author={Akari Asai and Zeqiu Wu and Yizhong Wang and Avirup Sil and Hannaneh Hajishirzi},
booktitle={Proceedings of ICLR},
year={2024},
url={https://openreview.net/forum?id=hSyW5go0v8}
}

@inproceedings{cuconasu2024power,
 author = {Cuconasu, Florin and Trappolini, Giovanni and Siciliano, Federico and Filice, Simone and Campagnano, Cesare and Maarek, Yoelle and Tonellotto, Nicola and Silvestri, Fabrizio},
 booktitle = {Proceedings of the 47th International ACM SIGIR Conference on Research and Development in Information Retrieval},
 pages = {719--729},
 title = {The power of noise: Redefining retrieval for rag systems},
 year = {2024},
 url = {https://arxiv.org/abs/2401.14887}
}

@inproceedings{shuster2021retrieval,
 author = {Shuster, Kurt and Poff, Spencer and Chen, Moya and Kiela, Douwe and Weston, Jason},
 booktitle = {Proceedings of EMNLP Findings},
 pages = {3784--3803},
 title = {Retrieval Augmentation Reduces Hallucination in Conversation},
 url = {https://aclanthology.org/2021.findings-emnlp.320.pdf},
 year = {2021}
}

@article{edge2024local,
 author = {Edge, Darren and Trinh, Ha and Cheng, Newman and Bradley, Joshua and Chao, Alex and Mody, Apurva and Truitt, Steven and Metropolitansky, Dasha and Ness, Robert Osazuwa and Larson, Jonathan},
 journal = {ArXiv preprint},
 title = {From local to global: A graph rag approach to query-focused summarization},
 url = {https://arxiv.org/abs/2404.16130},
 year = {2024}
}

@article{guo2024lightrag,
 author = {Guo, Zirui and Xia, Lianghao and Yu, Yanhua and Ao, Tu and Huang, Chao},
 journal = {ArXiv preprint},
 title = {Lightrag: Simple and fast retrieval-augmented generation},
 url = {https://arxiv.org/abs/2410.05779},
 year = {2024}
}

@article{vig2021exploring,
 author = {Vig, Jesse and Fabbri, Alexander R and Kry{\'s}ci{\'n}ski, Wojciech and Wu, Chien-Sheng and Liu, Wenhao},
 journal = {ArXiv preprint},
 title = {Exploring neural models for query-focused summarization},
 url = {https://arxiv.org/abs/2112.07637},
 year = {2021}
}

@article{jin2025search,
 author = {Jin, Bowen and Zeng, Hansi and Yue, Zhenrui and Yoon, Jinsung and Arik, Sercan and Wang, Dong and Zamani, Hamed and Han, Jiawei},
 journal = {ArXiv preprint},
 title = {Search-r1: Training llms to reason and leverage search engines with reinforcement learning},
 url = {https://arxiv.org/abs/2503.09516},
 year = {2025}
}

@article{liu2023lost,
 author = {Liu, Nelson F and Lin, Kevin and Hewitt, John and Paranjape, Ashwin and Bevilacqua, Michele and Petroni, Fabio and Liang, Percy},
 journal = {ArXiv preprint},
 title = {Lost in the middle: How language models use long contexts},
 url = {https://arxiv.org/abs/2307.03172},
 year = {2023}
}

@article{li2025search,
 author = {Li, Xiaoxi and Dong, Guanting and Jin, Jiajie and Zhang, Yuyao and Zhou, Yujia and Zhu, Yutao and Zhang, Peitian and Dou, Zhicheng},
 journal = {ArXiv preprint},
 title = {Search-o1: Agentic search-enhanced large reasoning models},
 url = {https://arxiv.org/abs/2501.05366},
 year = {2025}
}

@inproceedings{lewis2020retrieval,
 author = {Patrick S. H. Lewis and
Ethan Perez and
Aleksandra Piktus and
Fabio Petroni and
Vladimir Karpukhin and
Naman Goyal and
Heinrich K{\"{u}}ttler and
Mike Lewis and
Wen{-}tau Yih and
Tim Rockt{\"{a}}schel and
Sebastian Riedel and
Douwe Kiela},
 booktitle = {Proceedings of NeurIPS},
 title = {Retrieval-Augmented Generation for Knowledge-Intensive {NLP} Tasks},
 url = {https://proceedings.neurips.cc/paper/2020/hash/6b493230205f780e1bc26945df7481e5-Abstract.html},
 year = {2020}
}

@article{li2024structrag,
 author = {Li, Zhuoqun and Chen, Xuanang and Yu, Haiyang and Lin, Hongyu and Lu, Yaojie and Tang, Qiaoyu and Huang, Fei and Han, Xianpei and Sun, Le and Li, Yongbin},
 journal = {ArXiv preprint},
 title = {Structrag: Boosting knowledge intensive reasoning of llms via inference-time hybrid information structurization},
 url = {https://arxiv.org/abs/2410.08815},
 year = {2024}
}

@article{wu2025rankcot,
 author = {Wu, Mingyan and Liu, Zhenghao and Yan, Yukun and Li, Xinze and Yu, Shi and Zeng, Zheni and Gu, Yu and Yu, Ge},
 journal = {ArXiv preprint},
 title = {RankCoT: Refining Knowledge for Retrieval-Augmented Generation through Ranking Chain-of-Thoughts},
 url = {https://arxiv.org/abs/2502.17888},
 year = {2025}
}

@article{kuratov2024babilong,
 author = {Kuratov, Yury and Bulatov, Aydar and Anokhin, Petr and Rodkin, Ivan and Sorokin, Dmitry and Sorokin, Artyom and Burtsev, Mikhail},
 journal = {Proceedings of NeurIPS},
 pages = {106519--106554},
 title = {Babilong: Testing the limits of llms with long context reasoning-in-a-haystack},
 year = {2024},
 url = {https://proceedings.neurips.cc/paper_files/paper/2024/file/c0d62e70dbc659cc9bd44cbcf1cb652f-Paper-Datasets_and_Benchmarks_Track.pdf}
}

@article{li2023sequence,
 author = {Li, Tong and Wang, Zhihao and Shao, Liangying and Zheng, Xuling and Wang, Xiaoli and Su, Jinsong},
 journal = {ArXiv preprint},
 title = {A sequence-to-sequence\&set model for text-to-table generation},
 url = {https://arxiv.org/abs/2306.00137},
 year = {2023}
}

@article{jain2024structsum,
 author = {Jain, Parag and Marzoca, Andreea and Piccinno, Francesco},
 journal = {ArXiv preprint},
 title = {Structsum generation for faster text comprehension},
 url = {https://arxiv.org/abs/2401.06837},
 year = {2024}
}

@inproceedings{trivedi2023interleaving,
 author = {Trivedi, Harsh  and
Balasubramanian, Niranjan  and
Khot, Tushar  and
Sabharwal, Ashish},
 booktitle = {Proceedings of ACL},
 pages = {10014--10037},
 title = {Interleaving Retrieval with Chain-of-Thought Reasoning for Knowledge-Intensive Multi-Step Questions},
 url = {https://aclanthology.org/2023.acl-long.557},
 year = {2023}
}

@inproceedings{wang2025deepnotenotecentricdeepretrievalaugmented,
 author = {Ruobing Wang and Qingfei Zhao and Yukun Yan and Daren Zha and Yuxuan Chen and Shi Yu and Zhenghao Liu and Yixuan Wang and Shuo Wang and Xu Han and Zhiyuan Liu and Maosong Sun},
 booktitle = {Proceedings of EMNLP findings},
 title = {DeepNote: Note-Centric Deep Retrieval-Augmented Generation},
 url = {https://aclanthology.org/2025.findings-emnlp.1073/},
 year = {2025}
}

@article{klyne2004resource,
 author = {Klyne, Graham},
 journal = {http://www. w3. org/TR/rdf-concepts/},
 title = {Resource description framework (RDF): Concepts and abstract syntax},
 year = {2004},
 url = {https://www.w3.org/TR/2004/REC-rdf-concepts-20040210/}
}

@inproceedings{DBLP:conf/icml/GutierrezSQZ025,
  author       = {Bernal Jim{\'{e}}nez Guti{\'{e}}rrez and
                  Yiheng Shu and
                  Weijian Qi and
                  Sizhe Zhou and
                  Yu Su},
  editor       = {Aarti Singh and
                  Maryam Fazel and
                  Daniel Hsu and
                  Simon Lacoste{-}Julien and
                  Felix Berkenkamp and
                  Tegan Maharaj and
                  Kiri Wagstaff and
                  Jerry Zhu},
  title        = {From {RAG} to Memory: Non-Parametric Continual Learning for Large
                  Language Models},
  booktitle    = {Forty-second International Conference on Machine Learning, {ICML}
                  2025, Vancouver, BC, Canada, July 13-19, 2025},
  series       = {Proceedings of Machine Learning Research},
  publisher    = {{PMLR} / OpenReview.net},
  year         = {2025},
  url          = {https://proceedings.mlr.press/v267/gutierrez25a.html},
  bibsource    = {dblp computer science bibliography, https://dblp.org}
}

@inproceedings{DBLP:conf/emnlp/JiangXGSLDYCN23,
  author       = {Zhengbao Jiang and
                  Frank F. Xu and
                  Luyu Gao and
                  Zhiqing Sun and
                  Qian Liu and
                  Jane Dwivedi{-}Yu and
                  Yiming Yang and
                  Jamie Callan and
                  Graham Neubig},
  editor       = {Houda Bouamor and
                  Juan Pino and
                  Kalika Bali},
  title        = {Active Retrieval Augmented Generation},
  booktitle    = {Proceedings of the 2023 Conference on Empirical Methods in Natural
                  Language Processing, {EMNLP} 2023, Singapore, December 6-10, 2023},
  pages        = {7969--7992},
  publisher    = {Association for Computational Linguistics},
  year         = {2023},
  url          = {https://doi.org/10.18653/v1/2023.emnlp-main.495},
  doi          = {10.18653/V1/2023.EMNLP-MAIN.495},
  bibsource    = {dblp computer science bibliography, https://dblp.org}
}

@inproceedings{DBLP:conf/iclr/JinY0A25,
  author       = {Bowen Jin and
                  Jinsung Yoon and
                  Jiawei Han and
                  Sercan {\"{O}}. Arik},
  title        = {Long-Context LLMs Meet {RAG:} Overcoming Challenges for Long Inputs
                  in {RAG}},
  booktitle    = {The Thirteenth International Conference on Learning Representations,
                  {ICLR} 2025, Singapore, April 24-28, 2025},
  publisher    = {OpenReview.net},
  year         = {2025},
  url          = {https://openreview.net/forum?id=oU3tpaR8fm},
  bibsource    = {dblp computer science bibliography, https://dblp.org}
}

@inproceedings{DBLP:conf/acl/SuTA0024,
  author       = {Weihang Su and
                  Yichen Tang and
                  Qingyao Ai and
                  Zhijing Wu and
                  Yiqun Liu},
  editor       = {Lun{-}Wei Ku and
                  Andre Martins and
                  Vivek Srikumar},
  title        = {{DRAGIN:} Dynamic Retrieval Augmented Generation based on the Real-time
                  Information Needs of Large Language Models},
  booktitle    = {Proceedings of the 62nd Annual Meeting of the Association for Computational
                  Linguistics (Volume 1: Long Papers), {ACL} 2024, Bangkok, Thailand,
                  August 11-16, 2024},
  pages        = {12991--13013},
  publisher    = {Association for Computational Linguistics},
  year         = {2024},
  url          = {https://doi.org/10.18653/v1/2024.acl-long.702},
  doi          = {10.18653/V1/2024.ACL-LONG.702},
  bibsource    = {dblp computer science bibliography, https://dblp.org}
}

@inproceedings{DBLP:conf/acl/FangMM25,
  author       = {Jinyuan Fang and
                  Zaiqiao Meng and
                  Craig MacDonald},
  editor       = {Wanxiang Che and
                  Joyce Nabende and
                  Ekaterina Shutova and
                  Mohammad Taher Pilehvar},
  title        = {KiRAG: Knowledge-Driven Iterative Retriever for Enhancing Retrieval-Augmented
                  Generation},
  booktitle    = {Proceedings of the 63rd Annual Meeting of the Association for Computational
                  Linguistics (Volume 1: Long Papers), {ACL} 2025, Vienna, Austria,
                  July 27 - August 1, 2025},
  pages        = {18969--18985},
  publisher    = {Association for Computational Linguistics},
  year         = {2025},
  url          = {https://aclanthology.org/2025.acl-long.929/},
  bibsource    = {dblp computer science bibliography, https://dblp.org}
}

\clearpage
\appendix
\section{Appendix}
\subsection{License}
This section summarizes the licenses of the datasets used in our experiments.

All datasets used in this work permit academic use under their respective licenses and agreements: MuSiQue and HotpotQA are released under the CC-BY-4.0 License; 2WikiMQA is distributed under the Apache 2.0 License; and Bamboogle is released under the MIT License.



\subsection{Attention Visualization in \method{} During Iterative Retrieval}\label{appendix:attention}
\begin{figure}[t]
    \centering
    \subfigure[Attention Proportions over Indices and Documents.] { \label{fig:attention_index_passage} 
    \includegraphics[width=0.48\linewidth]{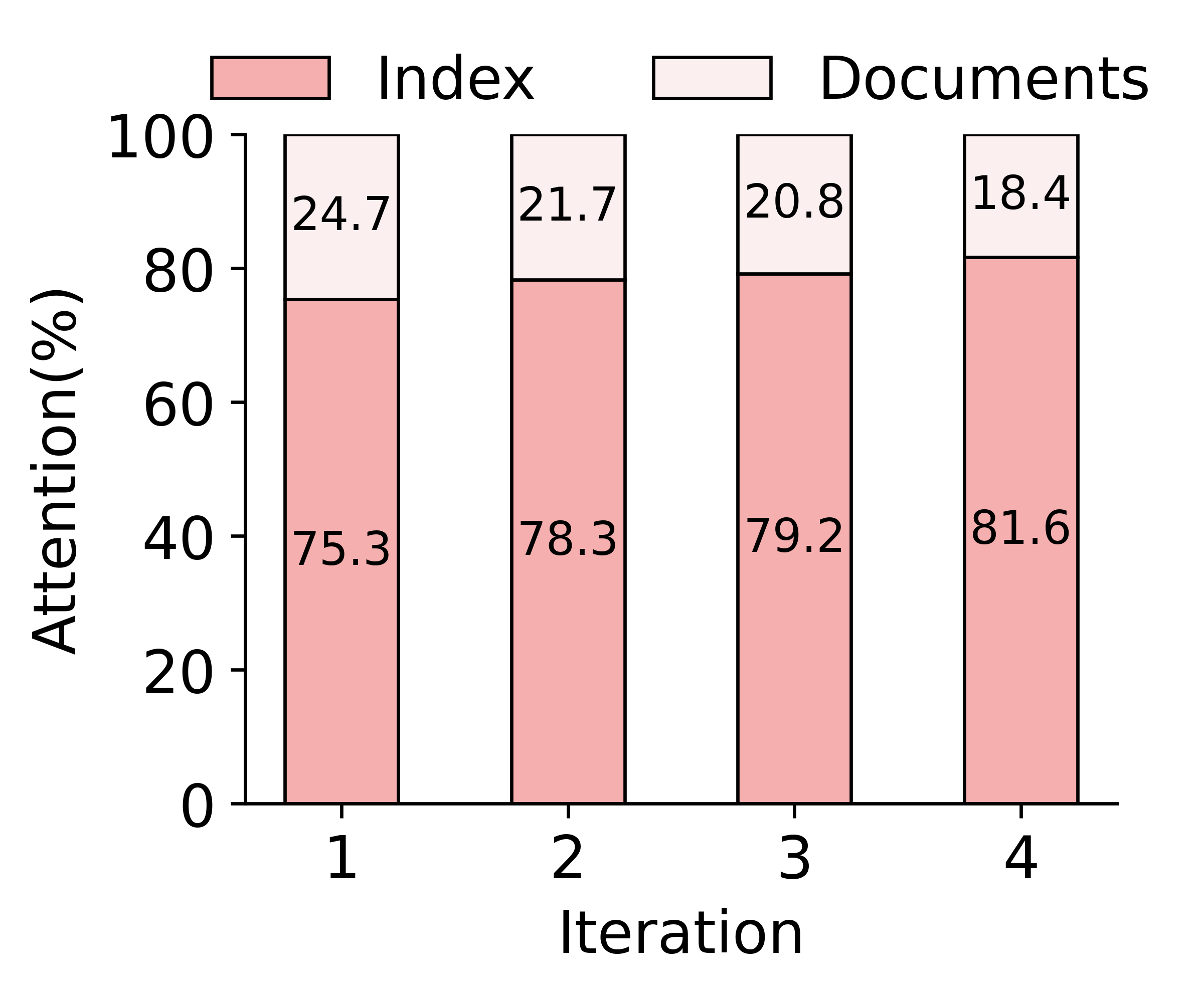}}
     \subfigure[Attention Entropy over Document Tokens.] { \label{fig:attention_entropy} 
    \includegraphics[width=0.46\linewidth]{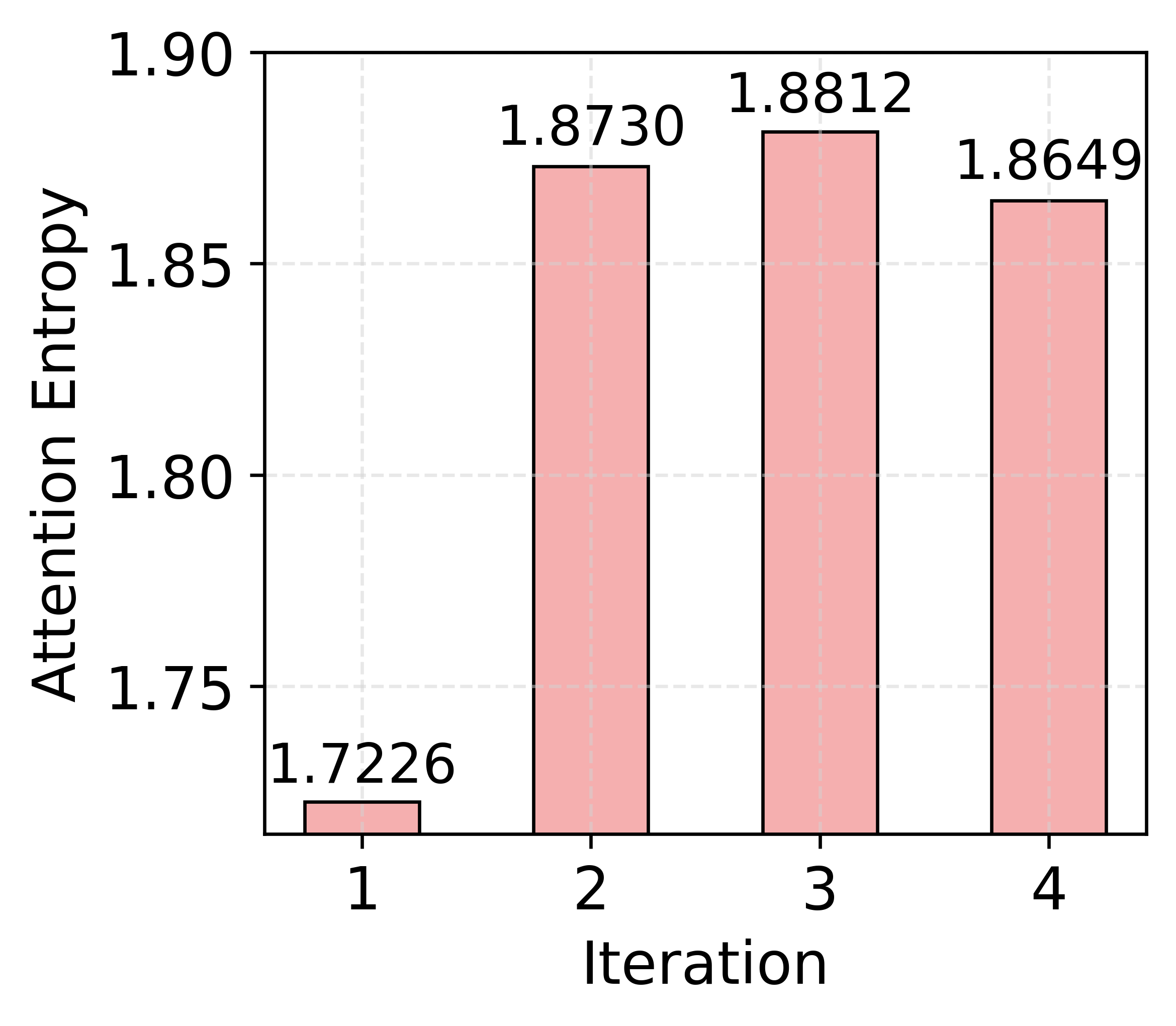}}
    \caption{Attention Distribution when Answering Queries with Documents and Indices in \method{}.}
    \label{fig:concentrated_attention}
\end{figure}

In this subsection, we explore how the constructed index assists LLMs in anchoring knowledge through the attention mechanism. Instances from HotpotQA that require the maximum number of retrieval steps (4 steps) are selected to illustrate the knowledge anchoring mechanism. As shown in Figure~\ref{fig:concentrated_attention}, we aggregate all retrieved documents as context and visualize the attention distribution on document tokens with indices constructed at different retrieval steps.

First, we show the attention proportions of both retrieved documents and the index during answer generation in Figure~\ref{fig:attention_index_passage} to examine the role of the constructed index. The results indicate that the index typically receives the majority of attention from the LLM during question answering, and that attention weights increase as the index is updated through more retrieval iterations. This demonstrates that the index plays a crucial role in helping LLMs better answer queries, confirming its effectiveness in anchoring retrieved knowledge.
Furthermore, Figure~\ref{fig:attention_entropy} presents the attention entropy when using the index and the answer output as ``queries'' to document tokens. The results show that the attention entropy increases when indices constructed at deeper iteration steps are used. This indicates that indices built with more retrieval iterations enable LLMs to allocate attention more selectively to the more query-relevant document tokens, thereby highlighting the effectiveness of the index in anchoring more distributed evidence from retrieved documents.

\subsection{Other Indexing Formats Exploration}\label{appendix:other_format}
Table~\ref{table/overall} shows the RAG performance of \method{} using keyworks\&Relationships (K\&R) index. In this subsection, we extend our investigation to a wide range of indexing formats, including Text, Keywords, Facts, and the combination of keywords\&facts (K\&F). We also report the RAG performance and ablation study of \method{} using these different indexing formats. 

\textbf{Index Definition.} We define four other distinct variants of the index representation $I$:\\
1) Textual Summary Index: The index is represented as a concise natural language summary $S$ that captures the core narrative of retrieved knowledge. 
\begin{equation}
    I_{text} = S.
\end{equation}
2) Keyword Index: The index is defined as a set of salient keywords $\mathcal{K} = \{k_1, k_2, \dots, k_n\}$ extracted from retrieved knowledge. 
\begin{equation}
    I_{keywords} = \{k_1, k_2, \dots, k_n\}.
\end{equation}
3) Fact Index: The index consists of a collection of discrete factual statements $\mathcal{F} = \{f_1, f_2, \dots, f_m\}$ extracted directly from retrieved knowledge.
\begin{equation}
    I_{facts} = \{f_1, f_2, \dots, f_m\}.
\end{equation}
4) Keyword and Fact Index: This representation extends keywords by associating each with a set of supporting facts. Formally, $I_{K\&F} = \{(k_i, \mathcal{F}_i)\}_{i=1}^n$, where each keyword $k_i$ is mapped to a list of facts $\mathcal{F}_i = \{f_{i,1}, f_{i,2}, \dots\}$.
\begin{equation}
\begin{aligned}
    & I_{K\&F} = \{k_1: \{f_{1,1}, f_{1,2}, \dots\}, \\
    & \dots, k_n: \{f_{n,1}, f_{n,2}, \dots\}\}.
\end{aligned}
\end{equation}

\textbf{Overall Performance of \method{} using Other Indexing Formats.}
As presented in Table~\ref{table/overall_add}, the experimental results demonstrate a clear and consistent performance gain across all four multi-hop QA datasets. The results strongly suggest that the structural complexity and logical connectivity of the index directly dictate the model's ability to perform accurate multi-hop reasoning. Among different index formats, relying on text summaries consistently yields the lowest performance, indicating that feeding the LLM dense, unstructured text blocks fails to mitigate noise, making it difficult for the model to associate distributed evidence. Extracting discrete ``Keywords'' or atomic ``Facts'' provides a substantial performance boost over plain text. However, because they present information as isolated nodes without explicit connections, which challenges further utilization of knowledge. The K\&R approach achieves the best performance. By pairing salient keywords with the explicit logical edges between them (Relations), K\&R transforms disjointed evidence into a structured cognitive indexing.

\textbf{Ablation Study of \method{} using Other Indexing Formats.}\label{appendix:add_ablation}
As presented in Table~\ref{table/additional_ablation}, the results consistently demonstrate that, regardless of the specific format, the full \method{} framework significantly outperforms both the vanilla RAG and its ablation models across all datasets.

A key observation is that structured indexing (Facts and K\&F) generally exhibits superior stability and performance compared to the pure Text Index. This suggests that decomposing retrieved knowledge into logical facts or key relations provides a cleaner ``anchoring'' point for the model. Noteably, the ablation of the evolving mechanism (w/o Evolving) leads to a substantial performance degradation across all indexing formats. This pattern confirms that the strength of our approach is not tied to a specific data structure, but rather to the incremental accumulation of the knowledge index. The evolving index serves as a dynamic navigational guidance that progressively bridges the logical gaps between retrieved documents.

Furthermore, comparing \method{} (QA w/ Index) and \method{} (QA w/ Docs) reveals that while both outperform the baseline, \method{} utilizes the index to guide the knowledge anchoring of raw documents, which achieves the highest performance. This highlights the importance of knowledge anchoring between structured indexing and raw evidence, where the index acts as a navigational guide within the cluttered retrieved context.

\begin{table}[t]
\centering
\small
\resizebox{\linewidth}{!}{
\begin{tabular}{lcccc}
\toprule 
\textbf{Method} &\textbf{MuSiQue} & \textbf{HotpotQA} & \textbf{2WikiMQA}  & \textbf{Bamboogle}\\ 
\midrule
IRCoT &{302.66s} &{1104.53s} &{367.47s} &{1153.27s}\\
IterRetGen &{235.84s} &{979.55s} &{251.82s} &{1180.53s} \\
Search R1 &{582.69s} &{857.93s} &{544.36s} &{1088.87s} \\
DeepNote &{1084.91s} &{1673.35s} &{1042.29s} &{1819.60s}\\
\method{} &{1207.50s} &{1239.99s} &{986.81s} &{1616.68s} \\
\bottomrule
\end{tabular}}
\caption{\label{table/latency} Inference Time of Different Iterative RAG Models implementing with Qwen2.5-7B-Instruct. We randomly sample 100 instances for testing.}
\end{table}

\subsection{Additional Experimental Details}\label{appendix:add_exp}
In this section, we first describe the reproduction details of the baseline methods, then present the prompt templates used in our experiments. Then, we report statistics on the retrieval steps reached across different datasets. Finally, we report the inference time and token usage of different iterative RAG methods.

\textbf{Baseline Description and Reproduction.} 
GraphRAG, LightRAG, and HippoRAG2 construct graphs to identify related information across multiple chunks. IRCoT and Search-R1 perform reasoning over retrieved knowledge and guide the model to identify and retrieve missing information. Iter-RetGen treats the information generated by the LLM based on retrieved knowledge as a cue to fetch further relevant information. DeepNote dynamically refines the retrieved knowledge into organized notes to facilitate knowledge accumulation.

For the Vanilla RAG model, following previous work~\cite{ram2023context}, we provide 5 retrieved documents as context to answer each question. GraphRAG, LightRAG, and HIPPORAG2 are reproduced
using their public implementations and are evaluated under a unified
retrieval and generation setting of ours for fair comparison. Due to computational restrictions, we index the top 10 documents for each question retrieved using bge-large-en-v1.5 rather than index the whole corpus. And then retrieve the top 5 documents from the built graph corpus for RAG generation.
We reproduce the IRCoT and Iter-RetGen baselines using FlashRAG~\cite{FlashRAG}. Search-R1 and DeepNote are reproduced based on their official open-source implementations.

\begin{figure*}[t] 
\centering
    \small
    \includegraphics[width=0.8\textwidth]{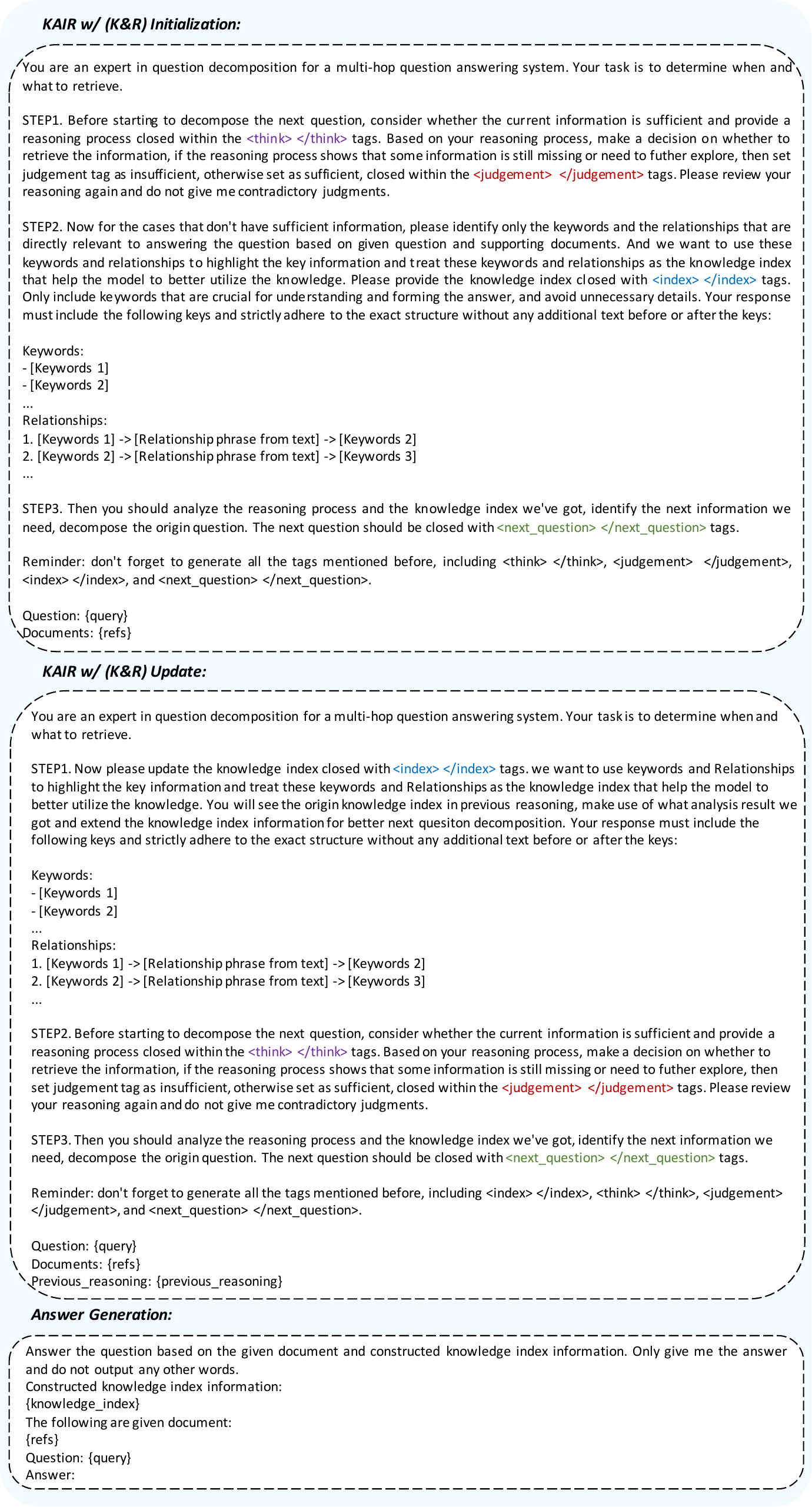}
    \caption{Prompt Templates Used for Implementing \method{} w/ (K\&R).} \label{fig:prompt_kr}
\end{figure*}

\begin{figure*}[t] 
\centering
    \includegraphics[width=0.8\textwidth]{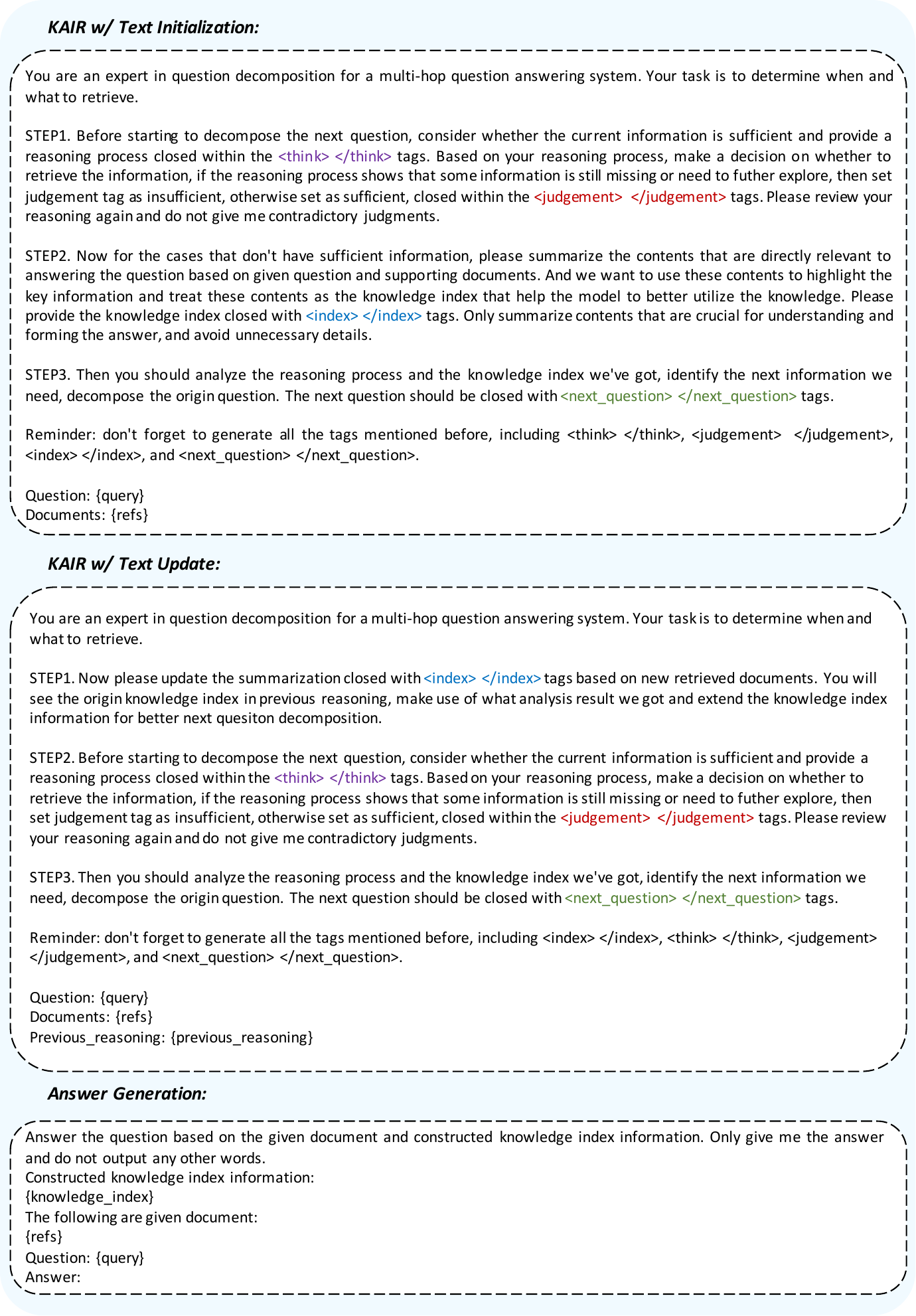}
    \caption{Prompt Templates Used for Implementing \method w/ Text.} \label{fig:prompt_text}
\end{figure*}

\begin{figure*}[t] 
\centering
    \includegraphics[width=0.8\textwidth]{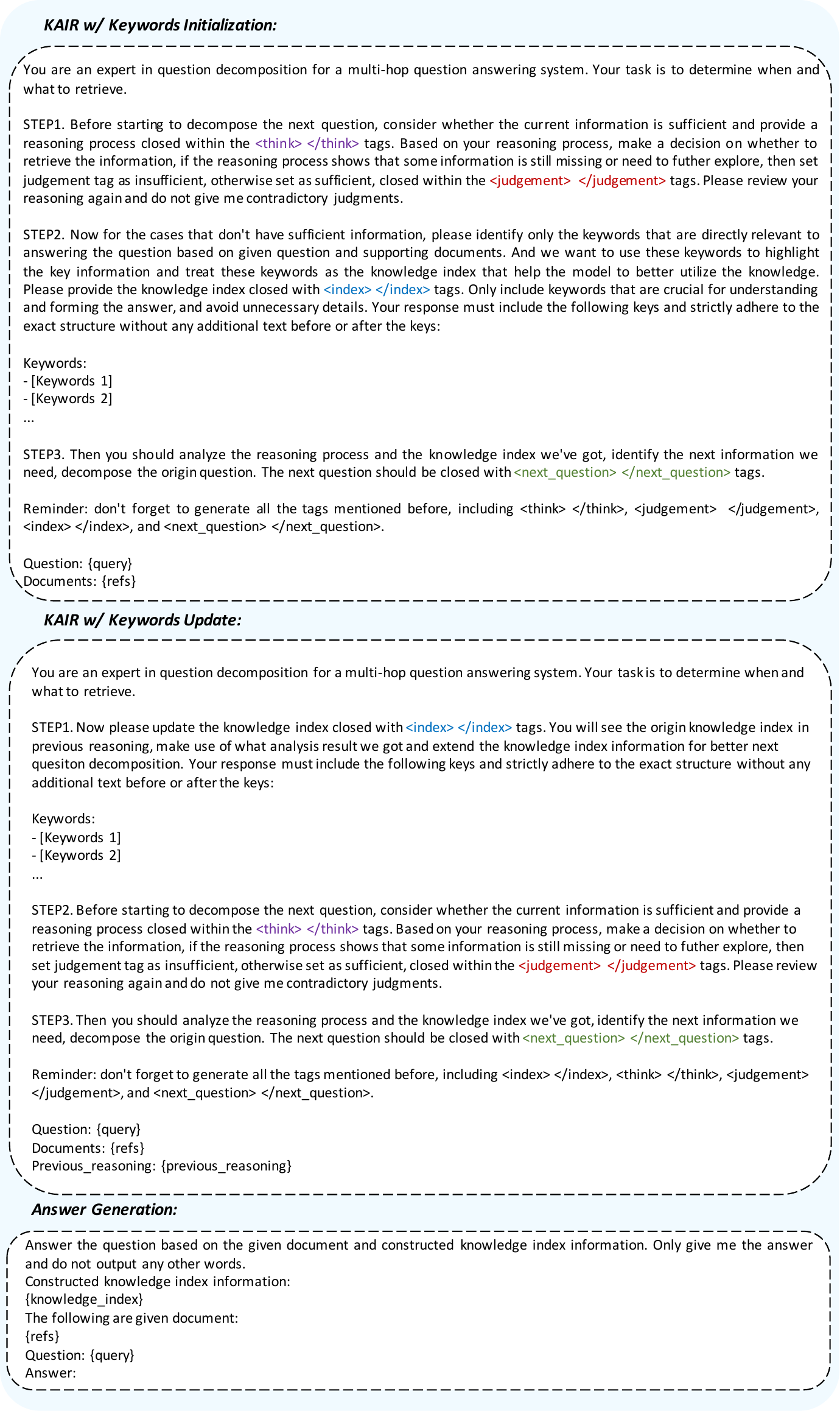}
    \caption{Prompt Templates Used for Implementing \method{} w/ Keywords.} \label{fig:prompt_keyword}
\end{figure*}

\begin{figure*}[t] 
\centering
    \includegraphics[width=0.8\textwidth]{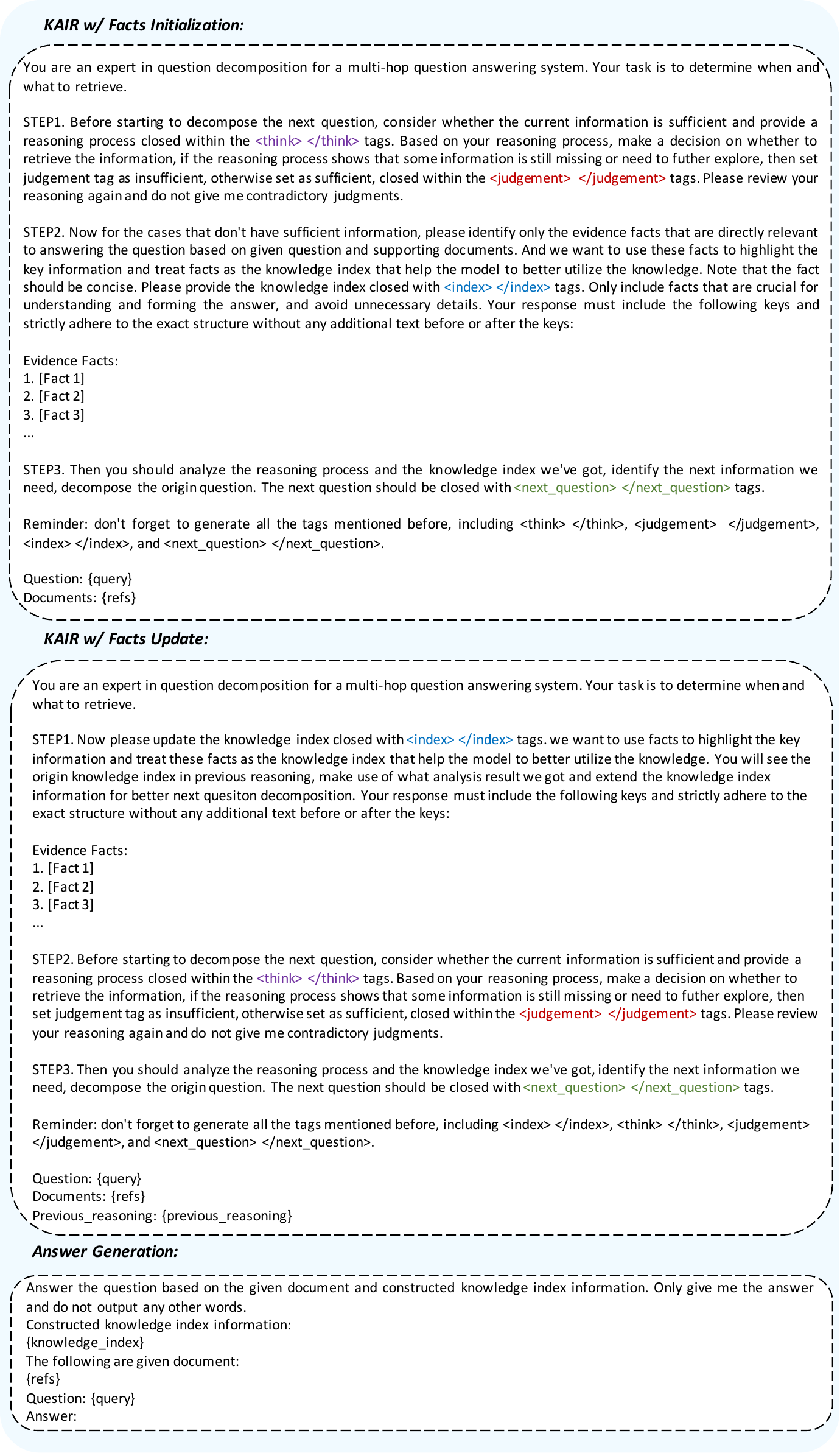}
    \caption{Prompt Templates Used for Implementing \method{} w/ Facts.} \label{fig:prompt_fact}
\end{figure*}

\begin{figure*}[t] 
\centering
    \includegraphics[width=0.7\textwidth]{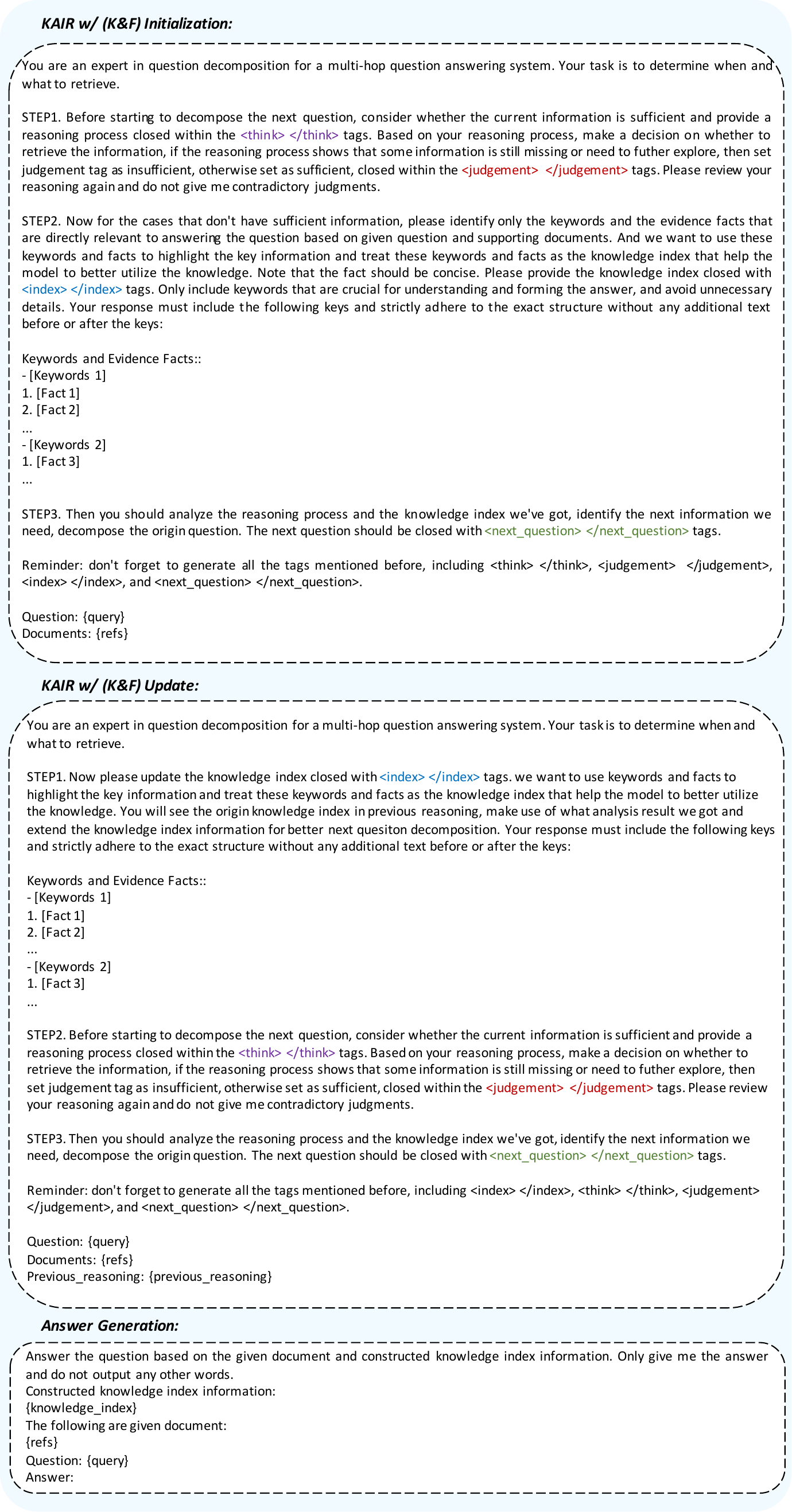}
    \caption{Prompt Templates Used for Implementing \method{} w/ (K\&F).} \label{fig:prompt_kf}
\end{figure*}

\textbf{Prompt Templates.} The instruction prompts used in \method{} w/ (K\&R) are shown in Figure~\ref{fig:prompt_kr}. During the indexing process, \method{} w/ (K\&R) initialization is applied at the first retrieval step, while \method{} w/ (K\&R) update is used in all subsequent steps. Similarly, the instruction prompts used in \method{} using different index formats, including text, keywords, facts, and (K\&F), are shown in Figure~\ref{fig:prompt_text}, Figure~\ref{fig:prompt_keyword}, Figure~\ref{fig:prompt_fact}, and Figure~\ref{fig:prompt_kf}, respectively.

\textbf{Dataset Statistics on Reached Retrieval Step.} Table~\ref{table/early_stop_all} reports dataset-level proportion statistics on the reached retrieval steps using \method{} w/ (K\&R), implemented with Qwen2.5-7B-Instruct and Llama3.1-8B-Instruct. Each dataset contains a total of 500 samples, except that Bamboogle contains a total of 125 samples. For both Qwen2.5-7B-Instruct and Llama3.1-8B-Instruct, the majority of cases terminate either at the second step or at the final fourth step, while relatively few queries stop at Step 3. Notably, Llama3.1-8B-Instruct shows a strong tendency to exhaust the full retrieval budget, with most samples reaching the maximum step across all datasets. In contrast, Qwen2.5-7B-Instruct exhibits greater decisiveness in earlier stages, frequently terminating at Step 2, particularly on datasets such as 2WikiMQA and Bamboogle.

\textbf{Inference Time of Different Iterative RAG Methods.}\label{appendix:inference_time}
This section reports the inference time of different iterative RAG methods implemented with Qwen2.5-7B-Instruct, as summarized in Table~\ref{table/latency}. Overall, methods that rely on deeper iterative reasoning or explicit structure construction incur higher inference costs. Lightweight approaches, such as IRCoT and IterRetGen, achieve relatively lower latency, whereas more expressive methods, including DeepNote, Search R1, and \method{}, exhibit increased inference time due to additional stages of information extraction, index construction, and structured reasoning. We further observe that inference time on the HotpotQA and Bamboogle datasets is consistently higher across all methods, reflecting the greater complexity of multi-hop reasoning and evidence aggregation required by these benchmarks.

\textbf{Token Usage of Different Iterative RAG Models.}\label{appendix:token_usage} We performed a detailed token analysis using the Qwen2.5-7B tokenizer across four multi-hop benchmarks.
While \method{} involves multiple reasoning steps, the core of our knowledge anchoring is inherently token-efficient. As shown in Table~\ref{table:token_usage_all}, the average output per reasoning step is consistently between 200 and 280 tokens. This structured format is far more compact than raw text summarization or exhaustive Chain-of-Thought (CoT) traces, serving as a high-density "semantic map" that guides the LLM without overwhelming the context window. And \method{} is not a "fixed-step" system that exhaustively runs all iterations. Through the knowledge sufficiency judgment mechanism, the system evaluates whether enough information has been anchored to answer the query. Our dataset statistics reveal that a significant portion of queries resolve in the earliest steps (as shown in Table~\ref{table/early_stop_all}). This "early exit" strategy drastically reduces the average cumulative cost per question, making it competitive with single-pass RAG systems that utilize large top-k retrieval.

\subsection{Performance of Vanilla RAG using Different Retrieval Top-k Documents}\label{appendix:vanilla}
In this section, we evaluate the performance of Vanilla RAG using different retrieval top-$k$ documents.

Table~\ref{table/vanilla} presents the performance comparison between \method{} and Vanilla RAG with varying retrieval top-$k$ documents. Our method consistently outperforms Vanilla RAG across all datasets and settings. Notably, simply increasing the number of retrieved documents may yield performance degradation. This indicates that one-pass retrieval struggles to handle the complexity of multi-hop questions. Expanding the context only introduces noise and distracts the LLM. In contrast, \method{} achieves substantial improvements by actively indexing evidence through iterations.

\subsection{Case Study}\label{appendix:case_study}
\begin{figure*}[t] 
\centering
    \includegraphics[width=1.0\textwidth]{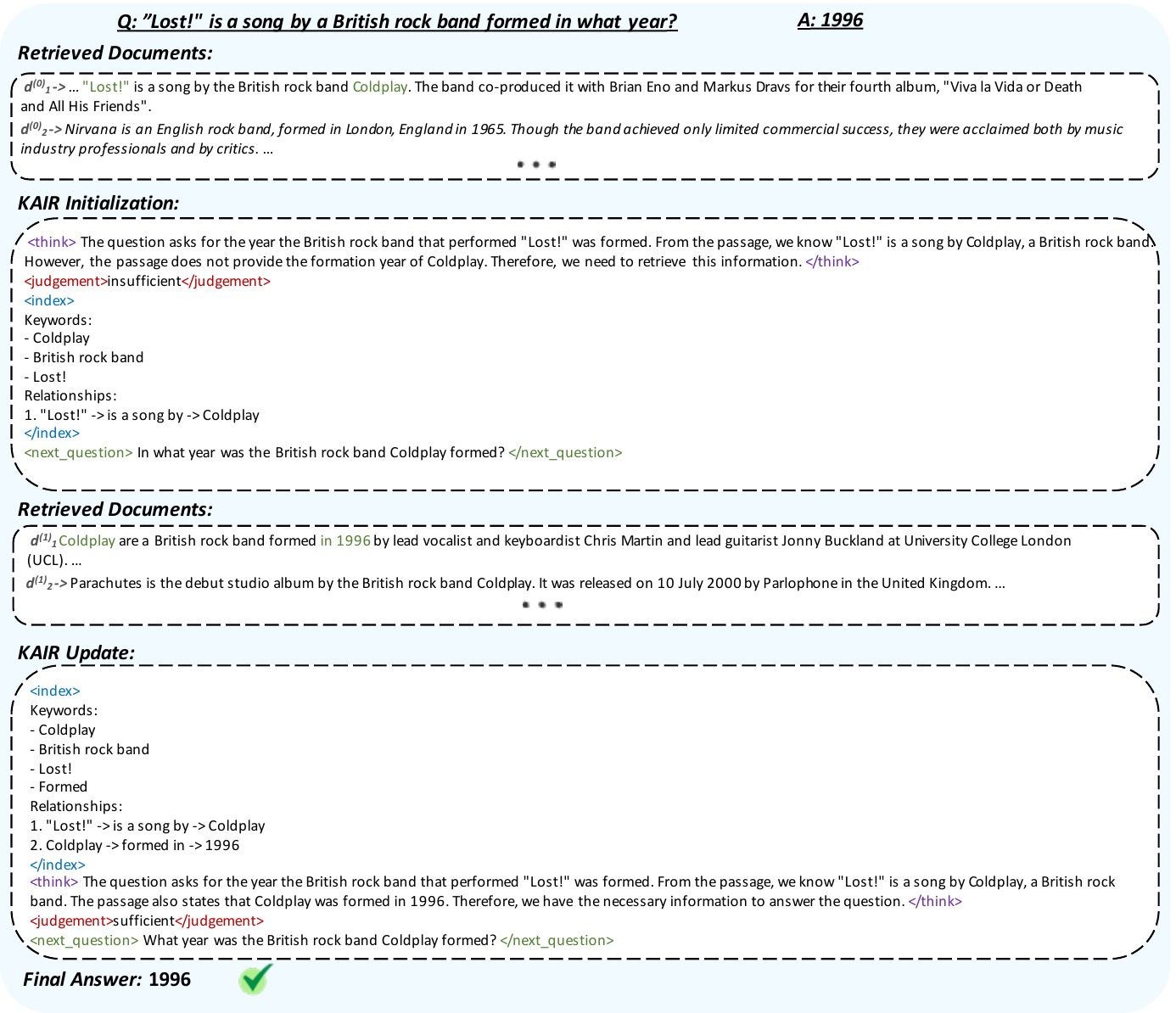}
    \caption{Case Study.} \label{fig:case_study}
\end{figure*}

Figure~\ref{fig:case_study} presents a case study illustrating how an anchoring index facilitates complex multi-hop reasoning.
The query asks for `` 'Lost!' is a song by a British rock band formed in what year'', in the initial stage, the model retrieves documents describing the song ``Lost!''. While these documents identify the band Coldplay, they lack the specific formation year required to answer the complex query. \method{} explicitly anchors the extracted keyword-relationship: ``Lost!'' -> is a song by -> Coldplay. By formalizing this link within the <index> tags, the model creates a structural waypoint that prevents the loss of intermediate reasoning results. Upon retrieving new documents, which mention Coldplay’s formation in 1996, the \method{} does not merely append new information, it integrates them into the existing structure. The relationship list is updated to bridge the gap: Coldplay -> formed in -> 1996. This explicit mapping transforms the reasoning task from a difficult global text search into a straightforward index-based lookup. This case study demonstrates that our method effectively maintains an anchoring index of intermediate reasoning states while actively resolving missing logical links through iterative retrieval and index updates.

\begin{table*}[ht]
\centering
\small
\begin{tabular}{lcccccccccc}
\toprule 
\multirow{2}{*}{\textbf{Method}} & \multicolumn{2}{c}{\textbf{MuSiQue}} & \multicolumn{2}{c}{\textbf{HotpotQA}} & \multicolumn{2}{c}{\textbf{2WikiMQA}} &\multicolumn{2}{c}{\textbf{Bamboogle}} &\multicolumn{2}{c}{\textbf{Average}}\\ 
&\textbf{F1} & \textbf{EM} & \textbf{F1}  & \textbf{EM}  & \textbf{F1} & \textbf{EM}  & \textbf{F1} & \textbf{EM} & \textbf{F1} & \textbf{EM} \\ 
\midrule
\multicolumn{11}{l}{\textit{Qwen2.5-7B-Instruct}}\\
\hdashline
Vanilla RAG~\shortcite{ram2023context} &{18.05} &{9.80} &{50.93} &{40.40} &{38.54} &{33.60} &{21.62} &{13.60} &{32.29} &{24.35} \\
IRCoT~\shortcite{trivedi2023interleaving} &{17.91} &{8.20} &{49.55} &{36.60} &{43.47} &{32.80} &{21.15} &{12.80} &{33.02} &{22.60} \\
Iter-RetGen~\shortcite{shao2023enhancing} &{21.36} &{13.00} &{58.81} &{47.40} &{43.15} &{37.60} &{24.82} &{16.80} &{37.04} &{28.70}  \\
Search-R1~\shortcite{jin2025search} &{24.95} &{17.60} &{54.78} &{43.40} &{49.55} &{40.60} &{25.13} &{20.00} &{38.60} &{30.40}  \\
DeepNote~\shortcite{wang2025deepnotenotecentricdeepretrievalaugmented} &{26.67} &{15.80} &{59.97} &{48.60} &{53.72} &{43.40} &{32.66} &{24.00} &{43.26} &{32.95} \\
\hdashline
\rowcolor{gray!8} \method{} w/ Text &{29.19} &{18.06} &{64.12} &{52.07} &{59.77} &{48.87} &{34.14} &{26.32} &{46.81} &{36.33}\\
\rowcolor{gray!8} w/ Keywords &{27.89} &{18.00} &\textbf{64.33} &\textbf{52.60} &{59.16} &{49.20} &{32.29} &{22.40} &{45.92} &{35.55}\\
\rowcolor{gray!8} w/ Facts &\textbf{31.22} &\textbf{21.29} &{64.04} &{51.90} &{59.45} &{50.20} &\textbf{36.32} &{25.60} &{47.76} &{37.25}\\
\rowcolor{gray!8} w/ (K\&F) &{30.63} &{20.84} &{64.08} &{52.00} &{59.93} &{50.80} &{31.96} &{22.40} &{46.65} &{36.51}\\
\rowcolor{gray!8} w/ (K\&R) &{31.17} &{21.20} &{62.64} &{50.80} &\textbf{60.99} &\textbf{52.21} &{36.28} &\textbf{27.20} &\textbf{47.77} &\textbf{37.85}\\
\midrule
\multicolumn{11}{l}{\textit{Llama3.1-8B-Instruct}}\\
\hdashline
Vanilla RAG~\shortcite{ram2023context} &{17.43} &{9.40} &{52.94} &{42.20} &{37.64} &{32.60} &{24.40} &{16.00} &{33.18} &{25.05}\\
IRCoT~\shortcite{trivedi2023interleaving} &{19.91} &{12.40} &{50.41} &{39.60} &{49.67} &{39.20} &{32.46} &{23.20} &{38.11} &{28.60}\\
Iter-RetGen~\shortcite{shao2023enhancing} &{18.15} &{11.20} &{55.97} &{43.80} &{34.32} &{26.60} &{26.09} &{20.00} &{33.63} &{25.40}\\
DeepNote~\shortcite{wang2025deepnotenotecentricdeepretrievalaugmented} &{26.81} &{16.60} &{59.51} &{47.80} &{56.65} &{45.80} &{37.46} &{27.20}&{45.11} &{34.35} \\
\hdashline
\rowcolor{gray!8} \method{} w/ Text &\textbf{34.10} &\textbf{24.14} &{65.75} &{51.45} &{62.17} &{52.21} &{36.18} &{27.27} &{49.55} &{38.77}\\
\rowcolor{gray!8} w/ Keywords &{28.00} &{16.90} &{62.77} &{49.90} &{55.50} &{46.15} &{34.40} &{27.64} &{45.17} &{35.15}\\
\rowcolor{gray!8} w/ Facts &{33.40} &{23.89} &{68.11} &{55.01} &{63.76} &{54.26} &{34.82} &{25.00} &{50.02} &{39.54}\\
\rowcolor{gray!8} w/ (K\&F) &{29.74} &{20.06} &{67.11} &{53.27} &{60.03} &{52.52} &{34.29} &{27.27} &{47.79} &{38.28}\\
\rowcolor{gray!8} w/ (K\&R) &{34.05} &{21.56} &\textbf{68.34} &\textbf{54.91} &\textbf{63.45} &\textbf{56.07} &\textbf{41.35} &\textbf{30.91} &\textbf{51.80} &\textbf{40.86}\\
\bottomrule
\end{tabular}
\caption{\label{table/overall_add} Overall Performance of \method{} using Other Indexing Formats.}
\end{table*}

\begin{table*}[ht]
\centering
\small
\resizebox{\linewidth}{!}{
\begin{tabular}{lcccccccccc}
\toprule 
\multirow{2}{*}{\textbf{Method}} & \multicolumn{2}{c}{\textbf{MuSiQue}} & \multicolumn{2}{c}{\textbf{HotpotQA}} & \multicolumn{2}{c}{\textbf{2WikiMQA}} &\multicolumn{2}{c}{\textbf{Bamboogle}} &\multicolumn{2}{c}{\textbf{Average}}\\ 
  & \textbf{F1}  & \textbf{EM}  & \textbf{F1}  & \textbf{EM}  & \textbf{F1} & \textbf{EM}  & \textbf{F1} & \textbf{EM} & \textbf{F1} & \textbf{EM} \\ 
\midrule
\rowcolor{gray!8} \multicolumn{11}{l}{\textit{Qwen2.5-7B-Instruct}}\\
\hdashline
Vanilla RAG &{18.05} &{9.80} &{50.93} &{40.40} &{38.54} &{33.60} &{21.62} &{13.60} &{32.29} &{24.35} \\
\hdashline
\multicolumn{11}{l}{\textit{Text Index}} \\
\method{} (QA w/ Docs) &{24.18} &{13.76} &{60.70} &{49.17} &{51.05} &{41.68} &{29.00} &{21.05} &{41.23} &{31.42} \\
\method{} (QA w/ Index) &{24.51} &{15.48} &{61.07} &{49.59} &{50.81} &{41.48} &\textbf{35.43} &\textbf{27.19} &{42.96} &{33.44} \\
\method{} &\textbf{29.19} &\textbf{18.06} &\textbf{64.12} &\textbf{52.07} &\textbf{59.77} &\textbf{48.87} &{34.14} &{26.32} &\textbf{46.81} &\textbf{36.33}\\
w/o Evolving &{22.15} &{11.80} &{57.02} &{44.06} &{40.20} &{29.20} &{24.24} &{16.00} &{35.90} &{25.27} \\
\hdashline
\multicolumn{11}{l}{\textit{Keywords Index}} \\
\method{} (QA w/ Docs) &{24.50} &{13.60} &{57.89} &{46.60} &{52.65} &{44.60} &{28.37} &{20.80} &{40.85} &{31.40} \\
\method{} (QA w/ Index) &{17.70} &{9.40} &{51.60} &{38.80} &{39.48} &{31.80} &{31.88} &{22.40} &{35.17} &{25.60} \\
\method{} &\textbf{27.89} &\textbf{18.00} &\textbf{64.33} &\textbf{52.60} &\textbf{59.16} &\textbf{49.20} &\textbf{32.29} &\textbf{22.40} &\textbf{45.92} &\textbf{35.55}
\\
w/o Evolving &{22.69} &{11.60} &{57.48} &{44.87} &{41.08} &{30.40} &{24.88} &{15.20} &{36.53} &{25.52} \\
\hdashline
\multicolumn{11}{l}{\textit{Facts Index}} \\
\method{} (QA w/ Docs) &{26.26} &{15.06} &{60.79} &{48.30} &{52.79} &{44.40} &{29.61} &{20.80} &{42.36} &{32.14} \\
\method{} (QA w/ Index) &{25.13} &{17.07} &{58.59} &{46.09} &{56.09} &{47.80} &{33.56} &{25.60} &{43.34} &{34.14} \\
\method{} &\textbf{31.22} &\textbf{21.29} &\textbf{64.04} &\textbf{51.90} &\textbf{59.45} &\textbf{50.20} &\textbf{36.32} &\textbf{25.60} &\textbf{47.76} &\textbf{37.25}\\
w/o Evolving &{21.66} &{11.40} &{57.58} &{44.87} &{41.79} &{30.20} &{24.72} &{16.80} &{36.44} &{25.82} \\
\hdashline
\multicolumn{11}{l}{\textit{(K\&F) Index}} \\
\method{} (QA w/ Docs) &{26.16} &{15.43} &{60.97} &{49.60} &{51.47} &{43.60} &{30.28} &{22.40} &{42.22} &{32.76} \\
\method{} (QA w/ Index) &{27.24} &{19.04} &{61.83} &{50.00} &{58.31} &{49.60} &\textbf{35.17} &\textbf{27.20} &{45.64} &{36.46} \\
\method{} &\textbf{30.63} &\textbf{20.84} &\textbf{64.08} &\textbf{52.00} &\textbf{59.93} &\textbf{50.80} &{31.96} &{22.40} &\textbf{46.65} &\textbf{36.51}\\
w/o Evolving &{23.14} &{13.20} &{56.31} &{44.67} &{39.88} &{28.60} &{25.74} &{16.80} &{36.27} &{25.82}\\
\hdashline
w/o Index &{22.99} &{12.63} &{55.05} &{43.60} &{50.64} &{42.80} &{26.66} &{18.40} &{38.84} &{29.36} \\
\midrule
\rowcolor{gray!8} \multicolumn{11}{l}{\textit{Llama3.1-8B-Instruct}}\\
\hdashline
Vanilla RAG &{17.43} &{9.40} &{52.94} &{42.20} &{37.64} &{32.60} &{24.40} &{16.00} &{33.18} &{25.05}\\
\hdashline
\multicolumn{11}{l}{\textit{Text Index}} \\
\method{} (QA w/ Docs) &{26.19} &{16.32} &{64.30} &{51.45} &{59.24} &{50.91} &{33.84} &{26.36} &{45.89} &{36.26} \\
\method{} (QA w/ Index) &{30.03} &{20.23} &{61.58} &{48.32} &{47.89} &{40.87} &{36.53} &{30.00} &{44.01} &{34.86} \\
\method{} &\textbf{34.10} &\textbf{24.14} &\textbf{65.75} &\textbf{51.45} &\textbf{62.17} &\textbf{52.21} &{36.18} &{27.27} &\textbf{49.55} &\textbf{38.77}\\
w/o Evolving &{20.70} &{12.50} &{61.17} &{48.20} &{46.16} &{33.18} &\textbf{39.50} &\textbf{35.00} &{41.88} &{32.22} \\
\hdashline
\multicolumn{11}{l}{\textit{Keywords Index}} \\
\method{} (QA w/ Docs) &{24.12} &{14.66} &{62.29} &{49.90} &{45.86} &{38.67} &{30.41} &{22.76} &{40.67} &{31.50} \\
\method{} (QA w/ Index) &{21.00} &{12.22} &{49.40} &{38.54} &{43.64} &{35.34} &{31.34} &{21.95} &{36.35} &{27.01} \\
\method{} &\textbf{28.00} &\textbf{16.90} &\textbf{62.77} &\textbf{49.90} &\textbf{55.50} &\textbf{46.15} &{34.40} &{27.64} &\textbf{45.17} &\textbf{35.15}\\
w/o Evolving &{21.60} &{12.75} &{63.08} &{49.64} &{45.65} &{35.53} &\textbf{39.86} &\textbf{35.00} &{42.55} &{33.23} \\
\hdashline
\multicolumn{11}{l}{\textit{Facts Index}} \\
\method{} (QA w/ Docs) &{24.84} &{16.39} &{65.59} &{52.70} &{50.70} &{42.12} &{34.58} &{26.00} &{43.93} &{34.30} \\
\method{} (QA w/ Index) &{30.89} &{20.28} &{66.84} &{53.98} &{62.01} &{51.94} &{37.09} &{27.00} &{49.21} &{38.30} \\
\method{} &\textbf{33.40} &\textbf{23.89} &\textbf{68.11} &\textbf{55.01} &\textbf{63.76} &\textbf{54.26} &{34.82} &{25.00} &\textbf{50.02} &\textbf{39.54}
\\
w/o Evolving &{20.74} &{12.50} &{63.84} &{50.84} &{49.28} &{39.06} &\textbf{39.83} &\textbf{27.50} &{43.42} &{32.48} \\
\hdashline
\multicolumn{11}{l}{\textit{(K\&F) Index}} \\
\method{} (QA w/ Docs) &{25.35} &{15.27} &{64.19} &{51.26} &{49.86} &{42.97} &{33.03} &{24.24} &{43.11} &{33.44} \\
\method{} (QA w/ Index) &{24.46} &{16.17} &{60.98} &{48.99} &{55.20} &{46.95} &{31.37} &{24.24} &{43.00} &{34.09} \\
\method{} &\textbf{29.74} &\textbf{20.06} &\textbf{67.11} &\textbf{53.27} &\textbf{60.03} &\textbf{52.52} &{34.29} &{27.27} &\textbf{47.79} &\textbf{38.28}\\
w/o Evolving &{20.74} &{12.50} &{61.12} &{48.20} &{43.56} &{31.06} &\textbf{38.55} &\textbf{32.50} &{40.99} &{31.07} \\
\hdashline
w/o Index &{21.46} &{13.48} &{63.03} &{49.64} &{46.82} &{38.59} &{33.03} &{26.92} &{41.09} &{32.16} \\
\bottomrule
\end{tabular}}
\caption{\label{table/additional_ablation} Ablation Study of \method{} using Other Indexing Formats. }
\end{table*}

\begin{table*}[ht]
\centering
\small
\renewcommand{\arraystretch}{1.05}
\resizebox{\linewidth}{!}{
\begin{tabular}{lllrrrr}
\toprule
\textbf{Index Format} & \textbf{Model} & \textbf{Step} & \textbf{MuSiQue} & \textbf{HotpotQA} & \textbf{2WikiMQA} & \textbf{Bamboogle} \\ 
\midrule

\multirow{8}{*}{\textbf{Text}} 
& \multirow{4}{*}{Qwen2.5-7B-Instruct} 
  & 1 & 9.89\%  & 32.78\% & 15.61\% & 10.53\% \\
& & 2 & 18.49\% & 21.58\% & 30.39\% & 23.68\% \\
& & 3 & 5.59\%  & 4.98\%  & 11.70\% & 9.65\%  \\
& & 4 & 66.02\% & 40.66\% & 42.30\% & 56.14\% \\
\cdashline{2-7}
& \multirow{4}{*}{Llama3.1-8B-Instruct} 
  & 1 & 0.23\%  & 1.57\%  & 2.28\%  & 1.82\%  \\
& & 2 & 20.92\% & 31.77\% & 39.04\% & 30.91\% \\
& & 3 & 10.80\% & 15.66\% & 18.72\% & 11.82\% \\
& & 4 & 68.05\% & 51.01\% & 39.95\% & 55.45\% \\
\midrule

\multirow{8}{*}{\textbf{Keywords}} 
& \multirow{4}{*}{Qwen2.5-7B-Instruct} 
  & 1 & 8.20\%  & 31.60\% & 15.60\% & 11.20\% \\
& & 2 & 21.20\% & 19.80\% & 27.20\% & 24.80\% \\
& & 3 & 6.40\%  & 5.60\%  & 8.20\%  & 4.80\%  \\
& & 4 & 64.20\% & 43.00\% & 49.00\% & 59.20\% \\
\cdashline{2-7}
& \multirow{4}{*}{Llama3.1-8B-Instruct} 
  & 1 & 0.61\%  & 1.01\%  & 1.46\%  & 0.81\%  \\
& & 2 & 8.15\%  & 15.21\% & 13.51\% & 9.76\%  \\
& & 3 & 2.85\%  & 5.48\%  & 8.94\%  & 6.50\%  \\
& & 4 & 88.39\% & 78.30\% & 76.09\% & 82.93\% \\
\midrule

\multirow{8}{*}{\textbf{Facts}} 
& \multirow{4}{*}{Qwen2.5-7B-Instruct} 
  & 1 & 5.02\%  & 27.05\% & 14.60\% & 8.80\%  \\
& & 2 & 18.47\% & 21.44\% & 27.40\% & 22.40\% \\
& & 3 & 8.23\%  & 5.81\%  & 13.40\% & 11.20\% \\
& & 4 & 68.27\% & 45.69\% & 44.60\% & 57.60\% \\
\cdashline{2-7}
& \multirow{4}{*}{Llama3.1-8B-Instruct} 
  & 1 & 0.83\%  & 0.77\%  & 1.03\%  & 2.00\%  \\
& & 2 & 16.11\% & 18.51\% & 29.97\% & 21.00\% \\
& & 3 & 11.67\% & 15.17\% & 15.76\% & 12.00\% \\
& & 4 & 71.39\% & 65.55\% & 53.23\% & 65.00\% \\
\midrule

\multirow{8}{*}{\textbf{(K\&F)}} 
& \multirow{4}{*}{Qwen2.5-7B-Instruct} 
  & 1 & 6.61\%  & 28.60\% & 15.20\% & 12.00\% \\
& & 2 & 17.23\% & 23.40\% & 39.00\% & 21.60\% \\
& & 3 & 7.21\%  & 5.60\%  & 7.60\%  & 3.20\%  \\
& & 4 & 68.94\% & 42.40\% & 38.20\% & 63.20\% \\
\cdashline{2-7}
& \multirow{4}{*}{Llama3.1-8B-Instruct} 
  & 1 & 0.30\%  & 1.00\%  & 1.59\%  & 0.00\%  \\
& & 2 & 12.28\% & 23.12\% & 27.32\% & 14.14\% \\
& & 3 & 10.78\% & 12.06\% & 12.20\% & 12.12\% \\
& & 4 & 76.65\% & 63.82\% & 58.89\% & 73.74\% \\
\midrule

\multirow{8}{*}{\textbf{(K\&R)}} 
& \multirow{4}{*}{Qwen2.5-7B-Instruct} 
  & 1 & 6.40\%  & 29.20\% & 17.67\% & 10.40\% \\
& & 2 & 23.40\% & 23.80\% & 40.96\% & 27.20\% \\
& & 3 & 8.20\%  & 5.00\%  & 11.04\% & 9.60\%  \\
& & 4 & 62.00\% & 42.00\% & 30.32\% & 52.80\% \\
\cdashline{2-7}
& \multirow{4}{*}{Llama3.1-8B-Instruct} 
  & 1 & 0.60\%  & 1.45\%  & 5.02\%  & 1.82\%  \\
& & 2 & 17.37\% & 22.18\% & 24.27\% & 16.36\% \\
& & 3 & 5.99\%  & 11.27\% & 12.55\% & 3.64\%  \\
& & 4 & 76.05\% & 65.09\% & 58.16\% & 78.18\% \\

\bottomrule
\end{tabular}}
\caption{\label{table/early_stop_all} Dataset Statistics on Reached Retrieval Step.}
\end{table*}
\begin{table*}[t]
\centering
\small
\setlength{\tabcolsep}{3.5pt} 
\begin{tabular}{llcccc}
\toprule
\textbf{Index Format} & \textbf{Step Label} & \textbf{HotpotQA (In/Out)} & \textbf{2WikiMQA (In/Out)} & \textbf{MuSiQue (In/Out)} & \textbf{Bamboogle (In/Out)} \\ \midrule

\multirow{6}{*}{Text} 
 & Step 0 & 831.5 / 205.1 & 1067.1 / 205.1 & 1007.6 / 205.1 & 862.4 / 205.1 \\
 & Step 1 & 953.9 / 215.8 & 1218.7 / 215.8 & 1140.9 / 215.8 & 961.1 / 215.8 \\
 & Step 2 & 1166.0 / 233.3 & 1398.8 / 233.3 & 1346.2 / 233.3 & 1156.7 / 233.3 \\
 & Step 3 & 1372.8 / 258.4 & 1567.1 / 258.4 & 1549.6 / 258.4 & 1357.9 / 258.4 \\
 & Final Gen & 881.4 / 6.6 & 1545.1 / 4.6 & 1388.1 / 6.7 & 1101.9 / 4.6 \\
 \cmidrule{2-6}
 & \textbf{Total Avg} & \textbf{3988.07} & \textbf{5651.40} & \textbf{6115.38} & \textbf{4919.39} \\ \midrule
 
\multirow{6}{*}{Keywords} 
 & Step 0 & 845.3 / 206.3 & 1078.2 / 207.4 & 1021.4 / 205.1 & 881.2 / 205.6 \\
 & Step 1 & 918.4 / 214.7 & 1195.6 / 211.7 & 1113.5 / 207.7 & 925.3 / 212.7 \\
 & Step 2 & 1074.5 / 233.7 & 1341.6 / 238.2 & 1256.4 / 232.9 & 1089.4 / 233.5 \\
 & Step 3 & 1216.7 / 258.6 & 1458.5 / 274.8 & 1412.2 / 264.4 & 1215.0 / 263.3 \\
 & Final Gen & 845.2 / 8.7 & 1469.5 / 4.6 & 1327.4 / 6.6 & 1021.3 / 4.6 \\
 \cmidrule{2-6}
 & \textbf{Total Avg} & \textbf{3927.19} & \textbf{5666.50} & \textbf{5865.93} & \textbf{4731.51} \\ \midrule

\multirow{6}{*}{Facts} 
 & Step 0 & 905.6 / 205.9 & 1130.8 / 205.9 & 1084.1 / 205.1 & 948.7 / 205.6 \\
 & Step 1 & 1066.6 / 215.9 & 1315.0 / 215.9 & 1241.0 / 215.8 & 1066.4 / 215.9 \\
 & Step 2 & 1282.2 / 233.3 & 1517.2 / 233.3 & 1445.0 / 233.3 & 1270.0 / 233.2 \\
 & Step 3 & 1479.5 / 258.6 & 1678.1 / 258.6 & 1645.0 / 258.4 & 1485.3 / 258.4 \\
 & Final Gen & 902.2 / 8.8 & 1528.4 / 4.6 & 1413.2 / 6.5 & 1042.0 / 4.6 \\
 \cmidrule{2-6}
 & \textbf{Total Avg} & \textbf{4505.68} & \textbf{6020.12} & \textbf{6636.20} & \textbf{5282.06} \\ \midrule

\multirow{6}{*}{(K\&F)} 
 & Step 0 & 921.3 / 205.9 & 1150.6 / 205.9 & 1103.0 / 205.9 & 963.5 / 205.9 \\
 & Step 1 & 1076.8 / 215.9 & 1333.6 / 215.9 & 1264.5 / 215.9 & 1095.7 / 215.9 \\
 & Step 2 & 1310.1 / 233.3 & 1551.2 / 233.3 & 1487.7 / 233.3 & 1312.7 / 233.3 \\
 & Step 3 & 1540.2 / 258.6 & 1738.1 / 258.6 & 1713.7 / 258.4 & 1530.8 / 258.4 \\
 & Final Gen & 917.9 / 8.8 & 1526.2 / 4.7 & 1448.0 / 6.6 & 1082.1 / 4.7 \\
 \cmidrule{2-6}
 & \textbf{Total Avg} & \textbf{4453.73} & \textbf{5746.86} & \textbf{6775.41} & \textbf{5437.18} \\ \midrule

\multirow{6}{*}{(K\&R)} 
 & Step 0 & 920.5 / 205.9 & 1157.5 / 205.9 & 1094.3 / 205.9 & 958.1 / 205.9 \\
 & Step 1 & 1050.5 / 215.9 & 1341.7 / 215.9 & 1240.1 / 215.9 & 1075.4 / 215.9 \\
 & Step 2 & 1278.3 / 233.3 & 1537.4 / 233.3 & 1444.4 / 233.3 & 1262.2 / 233.3 \\
 & Step 3 & 1492.1 / 258.4 & 1745.1 / 258.4 & 1659.2 / 258.4 & 1490.4 / 258.4 \\
 & Final Gen & 891.0 / 8.7 & 1538.7 / 4.6 & 1386.7 / 6.6 & 1083.5 / 4.6 \\
 \cmidrule{2-6}
 & \textbf{Total Avg} & \textbf{4368.44} & \textbf{5495.90} & \textbf{6388.48} & \textbf{5142.23} \\ 

\bottomrule
\end{tabular}
\caption{Detailed Token Usage (Input / Output) per Step across all Methods and Datasets.``In'' represents prompt tokens, and ``Out'' represents generated reasoning/answer tokens.}
\label{table:token_usage_all}
\end{table*}

\begin{table*}[ht]
\centering
\small
\resizebox{\linewidth}{!}{
\begin{tabular}{lcccccccccc}
\toprule 
\multirow{2}{*}{\textbf{Method}} & \multicolumn{2}{c}{\textbf{MuSiQue}} & \multicolumn{2}{c}{\textbf{HotpotQA}} & \multicolumn{2}{c}{\textbf{2WikiMQA}} &\multicolumn{2}{c}{\textbf{Bamboogle}} &\multirow{2}{*}{\textbf{F1 Avg.}} &\multirow{2}{*}{\textbf{EM Avg.}}\\ 
&\textbf{F1} & \textbf{EM} & \textbf{F1}  & \textbf{EM}  & \textbf{F1} & \textbf{EM}  & \textbf{F1} & \textbf{EM} \\ 
\midrule
\multicolumn{11}{l}{\textit{Qwen2.5-7B-Instruct}}\\
\hdashline
Vanilla RAG (top5) &{18.05} &{9.80} &{50.93} &{40.40} &{38.54} &{33.60} &{21.62} &{13.60} &{32.29} &{24.35}\\
Vanilla RAG (top10) &{21.39} &{12.40} &{56.24} &{43.60} &{38.26} &{32.80} &{26.16} &{18.40} &{35.51} &{26.80}\\
Vanilla RAG (top15) &{20.29} &{11.20} &{56.29} &{44.60} &{40.98} &{35.00} &{27.12} &{20.00} &{36.17} &{27.70}\\
Vanilla RAG (top20) &{20.80} &{12.40} &{56.79} &{43.60} &{41.13} &{35.20} &{24.38} &{15.20} &{35.78} &{26.60}\\
\hdashline
\rowcolor{gray!8} \method{} w/ Text &{29.19} &{18.06} &{64.12} &{52.07} &{59.77} &{48.87} &{34.14} &{26.32} &{46.81} &{36.33}\\
\rowcolor{gray!8} w/ Keywords &{27.89} &{18.00} &\textbf{64.33} &\textbf{52.60} &{59.16} &{49.20} &{32.29} &{22.40} &{45.92} &{35.55}\\
\rowcolor{gray!8} w/ Facts &\textbf{31.22} &\textbf{21.29} &{64.04} &{51.90} &{59.45} &{50.20} &\textbf{36.32} &{25.60} &{47.76} &{37.25}\\
\rowcolor{gray!8} w/ (K\&F) &{30.63} &{20.84} &{64.08} &{52.00} &{59.93} &{50.80} &{31.96} &{22.40} &{46.65} &{36.51}\\
\rowcolor{gray!8} w/ (K\&R) &{31.17} &{21.20} &{62.64} &{50.80} &\textbf{60.99} &\textbf{52.21} &{36.28} &\textbf{27.20} &\textbf{47.77} &\textbf{37.85} \\
\midrule
\multicolumn{11}{l}{\textit{Llama3.1-8B-Instruct}}\\
\hdashline
Vanilla RAG (top5) &{17.43} &{9.40} &{52.94} &{42.20} &{37.94} &{32.60} &{24.40} &{16.00} &{33.18} &{25.05}\\
Vanilla RAG (top10) &{21.56} &{12.20} &{56.33} &{43.80} &{38.63} &{32.80} &{26.09} &{18.40} &{35.65} &{26.80}\\
Vanilla RAG  (top15) &{21.20} &{12.00} &{55.92} &{44.20} &{42.16} &{36.40} &{27.50} &{19.20} &{36.70} &{27.95}\\
Vanilla RAG (top20) &{20.68} &{11.80} &{57.03} &{43.40} &{39.76} &{33.80} &{26.64} &{18.40} &{36.03} &{26.85}\\
\hdashline
\rowcolor{gray!8} \method{} w/ Text &\textbf{34.10} &\textbf{24.14} &{65.75} &{51.45} &{62.17} &{52.21} &{36.18} &{27.27} &{49.55} &{38.77}\\
\rowcolor{gray!8} w/ Keywords &{28.00} &{16.90} &{62.77} &{49.90} &{55.50} &{46.15} &{34.40} &{27.64} &{45.17} &{35.15}\\
\rowcolor{gray!8} w/ Facts &{33.40} &{23.89} &{68.11} &{55.01} &{63.76} &{54.26} &{34.82} &{25.00} &{50.02} &{39.54}\\
\rowcolor{gray!8} w/ (K\&F) &{29.74} &{20.06} &{67.11} &{53.27} &{60.03} &{52.52} &{34.29} &{27.27} &{47.79} &{38.28}\\
\rowcolor{gray!8} w/ (K\&R) &{34.05} &{21.56} &\textbf{68.34} &\textbf{54.91} &\textbf{63.45} &\textbf{56.07} &\textbf{41.35} &\textbf{30.91} &\textbf{51.80} &\textbf{40.86} \\

\bottomrule
\end{tabular}}
\caption{\label{table/vanilla} Vanilla RAG using Different Retrieval Top-k Documents.}
\end{table*}


\end{document}